\documentclass[10pt,twocolumn,letterpaper]{article}

\usepackage[pagenumbers]{cvpr} 

\usepackage{graphicx}
\usepackage{amsmath}
\usepackage{amssymb}
\usepackage{booktabs}
\usepackage{times}
\usepackage{epsfig}
\usepackage{subfigure}
\usepackage{tabu}

\usepackage{multirow}
\usepackage[numbers]{natbib}
\usepackage{stfloats}
\usepackage{colortbl}

\newcommand{\boldblue}[1]{\textcolor{blue}{\textbf{#1}}}
\newcommand{\boldblack}[1]{\textcolor{black}{\textbf{#1}}}

\usepackage[pagebackref,breaklinks,colorlinks]{hyperref}

\usepackage[capitalize]{cleveref}
\crefname{section}{Sec.}{Secs.}
\Crefname{section}{Section}{Sections}
\Crefname{table}{Table}{Tables}
\crefname{table}{Tab.}{Tabs.}

\begin{document}


\title{Learning Holistic Geometric Representations for Monocular 3D Object Detection}

\author{Yinmin Zhang\textsuperscript{1}, \ \ 
	Xinzhu Ma\textsuperscript{2},\ \  
	Shuai Yi\textsuperscript{1}, \ \
	Jun Hou\textsuperscript{1}, \ \
	Zhihui Wang\textsuperscript{3},  \ \
	Wanli Ouyang\textsuperscript{2} \\  
	Dan Xu\textsuperscript{4}, \\	
\textsuperscript{1}SenseTime Research, \ \
\textsuperscript{2}The University of Sydney, \ \
\textsuperscript{3}Dalian University of Technology\\
\textsuperscript{4}The Hong Kong University of Science and Technology,\\
{\tt\small \{zhangyinmin, yishuai, houjun\}@sensetime.com,}
\\
{\tt\small \{xinzhu.ma, wanli.ouyang\}@sydney.edu.au,}
\\
{\tt\small zhwang@dlut.edu.au,}
{\tt\small danxu@cse.ust.hk}
}

\maketitle

\begin{abstract}
As a crucial task of autonomous driving, 3D object detection has made significant progress in recent years. 
However, monocular 3D object detection remains a challenging problem due to the unsatisfactory performance in depth estimation. 
Most existing monocular methods typically directly regress the depth, while ignoring essential relationships between the depth and various geometric elements (e.g.~bounding box sizes, 3D object dimensions, and object poses).
In this paper, we propose to learn geometry-guided depth estimation with projective modeling to advance monocular 3D object detection. Specifically, a principled geometry formula with projective modeling of 2D and 3D depth predictions in the monocular 3D object detection network is devised.
We further implement and embed the proposed formula to enable geometry-aware deep representation learning, allowing effective 2D and 3D interactions for boosting the depth estimation. 
Moreover, we provide a strong baseline through addressing substantial misalignment between 2D annotation and projected boxes to ensure robust learning with the proposed holistic geometric formula. 
Experiments on the KITTI dataset show that our method remarkably improves the detection performance of the state-of-the-art monocular-based method without extra data by \textbf{2.80\%} on the moderate test setting. The model and code will be released upon acceptance.
\end{abstract}
\vspace{-4pt}

\begin{figure}[t]
\begin{center}
\includegraphics[width=0.92\linewidth]{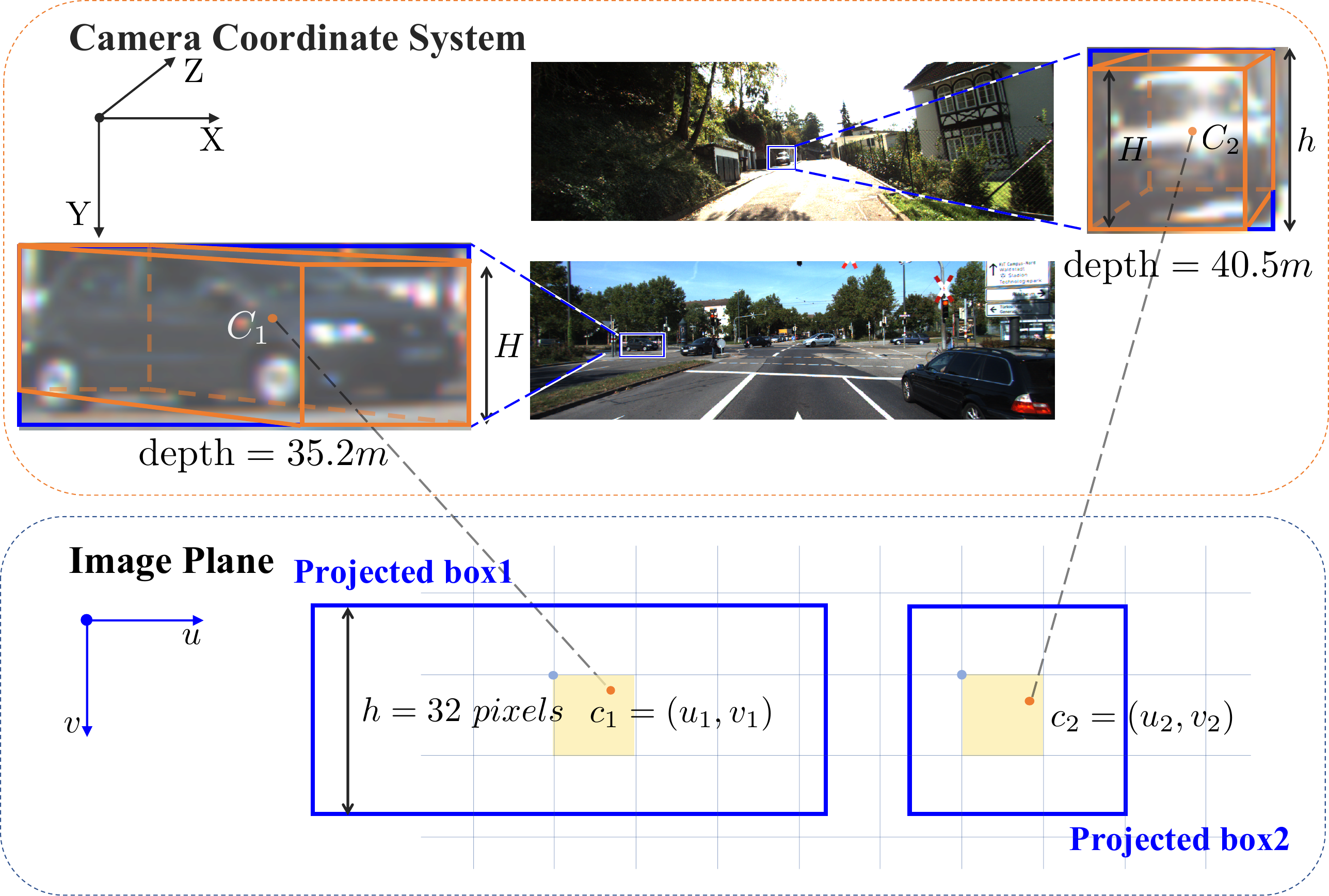}
\end{center}
\caption{Visualization of the depth difference in geometry projection. Object depth critically relates to both the \emph{pose} and \emph{position} of the object. For instance, for two cars with the same height in both the 2D bounding box (in blue) and the 3D bounding box (in orange), the depth values of their centers differ by more than $5$ meters because of their distinct \emph{poses} and \emph{positions}.
}
\label{fig: misalignment}
\end{figure}
\section{Introduction}
As an important and challenging problem, 3D object detection plays a fundamental role in various computer vision applications, such as autonomous driving, robotics, and augmented/virtual reality.
In recent years monocular 3D object detection has received great attention, because it simply uses a monocular camera instead of requiring extra sensing devices as in LiDAR-based~\cite{pv-rcnn, pointpillars, fast, shi2020points} and stereo-based~\cite{chen20153d, stereorcnn, fpoint, xu2018multi} methods.
However, the performance gap between LiDAR-based and monocular image-based approaches remains significant, mainly because of the lack of reliable depth information. A quantitative investigation is conducted by replacing the depth predictions with the ground-truth depth values on a baseline model. The detection performance can be remarkably improved from 11.84\% to 70.91\% in terms of the ${\rm AP}_{40}$ under the moderate setting of car category on the KITTI {\it val} set (see Table~\ref{tb: error}), which suggests that the depth estimation is a critical performance bottleneck in the monocular 3D object detection. 

The depth information has also been successfully applied as an important 3D geometry element to facilitate the learning in other problems, such as 2D object detection~\cite{reppoints, xu2019geometry, hoiem2008putting}, human pose estimation \cite{rhodin2018unsupervised}, and camera localization~\cite{XuMoving,brahmbhatt2018geometry, Sheng2019Unsupervised}. 
However, how to jointly model the geometry relationships between the depth and different 2D/3D network predictions, such as 2D box sizes, 3D dimensions, and poses, and enable joint learning with the modeled geometry constraints for geometry-aware monocular 3D detection is rarely explored in the literature.
An intuitive way to introduce the geometric relationships is to leverage perspective projection between the 3D scene space and the 2D image plane.~{Prior works~\cite{keypoint, decoupled, gs3d, rtm3d} 
either weakly use the geometry considering the projection consistency between 2D and 3D
for post-processing or employ perspective projection regardless of the object poses and positions. However, object poses and positions can provide considerably stronger geometric constraints and are extremely important for accurate depth estimation. As can be observed in Fig.~\ref{fig: misalignment}, the depth values differ by more than 5 meters due to the distinct poses and positions of the cars with the same height of 2D/3D boxes.}

In this paper, we propose an effective holistic geometric formula by principled modeling of the relationships between the depth and different geometry elements predicted from the deep network for the task of monocular 3D object detection, including 2D bounding boxes, 3D object dimensions, object poses, and object positions. We further implement the proposed formula to develop a geometry-based network module, which can be flexibly embedded into the deep learning framework, allowing effective geometry-aware learning on the representation level for guiding the depth estimation and advancing the monocular 3D object detection.  Besides, the geometry module can be utilized during both the training and inference phases without additional complex post-processing. 
Moreover, we provide a simple yet strong baseline for ensuring robust learning with the proposed geometry module, which is achieved through addressing the severe misalignment between the annotated 2D box and the projected 2D box from the 3D annotations.
This effective baseline achieves an AP of 13.37\% under the moderate setting of car category on the KITTI \emph{val} set.

\par To summarize, the contribution of this paper is threefold:

\begin{itemize}
\setlength{\itemsep}{1pt}
\setlength{\parsep}{1pt}
\setlength{\parskip}{1pt}
\item We propose an effective holistic geometric formula, which jointly models the perspective geometry relationships of multiple 2D/3D elements predicted from the deep neural network, providing strong geometric constraints for depth enhancement learning. 

\item We implement the proposed geometric formula in neural network as a module, which can be leveraged to guide the representation learning for boosting the depth estimation to significantly advance the performance of the monocular 3D object detection.

\item 
{
We provide a simple yet strong baseline through dealing with the misalignment between 2D projected boxes and 2D annotation boxes, which achieves
13.37\% on the moderate of the KITTI \emph{val} set.
We expect our baseline will be beneficial for the community in future research on monocular 3D object detection.
}
\end{itemize}

Extensive experiments conducted on the challenging KITTI~\cite{kitti} dataset clearly demonstrate the effectiveness of the proposed approach and show that our method achieves \textbf{13.81\%} in terms of the $\rm {AP}_{40}$ metric, which is \textbf{2.80\%} absolute $\rm {AP}_{40}$ improvement compared with the state-of-the-art monocular 3D object detection method on the moderate setting of the KITTI \textit{test} set for the car category.

\section{Related Work}
There are two groups of works closely related to ours, \ie ~monocular 3D object detection and geometry-guided 3D object detection.

\par\noindent\textbf{Monocular 3D Object Detection.}
Compared with the methods with LiDAR and stereo sensors, 3D object detection with monocular images is challenging due to the absence of reliable depth information. Existing works~\cite{deepmanta, roi10d, am3d, patchnet, decoupled, d4lcn} have considered using external pretrained networks, extra training data, and prior knowledge to improve the performance of monocular 3D object detection.
Particularly, DeepMANTA~\cite{deepmanta} utilizes extra 3D shape and template in learning 2D/3D vehicle models, and performs 2D/3D matching for the detection.~{Inspired by the importance of accurate depth for 3D object detection, many works~\cite{CaDDN, am3d, patchnet, d4lcn, DA-3Ddet} develop monocular 3D object detection by introducing pretrained external network for depth estimation. In contrast to these methods, we only use the monocular image as input without any extra burden.}
\par In recent years, some works also only use RGB data as the input for the task~\cite{monodis, m3drpn, monopair, MoVi-3D, liu2020reinforced}. 
For instance, MonoDIS~\cite{monodis} proposes to leverage a disentangling transformation between different 2D and 3D tasks to optimize the parameters at the loss level.
M3D-RPN~\cite{m3drpn} focuses on the design of depth-aware convolution layers to improve 3D parameter estimation and post-optimization of the orientation by exploring the consistency between projected and annotated bounding boxes. 
To address the common occlusion issue in monocular object detection, MonoPair~\cite{monopair} proposes to model spatial relationships of objects in paired adjacent RGB images via introducing an uncertainty-based prediction for improving the detection. {MoVi-3D~\cite{MoVi-3D} builds virtual views where the object appearance is normalized depending on the distance to reduce the visual appearance variability. RAR-Net~\cite{liu2020reinforced} builds a post-processing method by introducing reinforcement learning to improve the 3D object detection performance.}
Although these existing methods achieved very promising results, the beneficial geometry relationships between the different 2D and 3D predictions from the network are not explicitly modeled.

\begin{figure*}[!t]
\begin{center}
\includegraphics[width=0.97\linewidth]{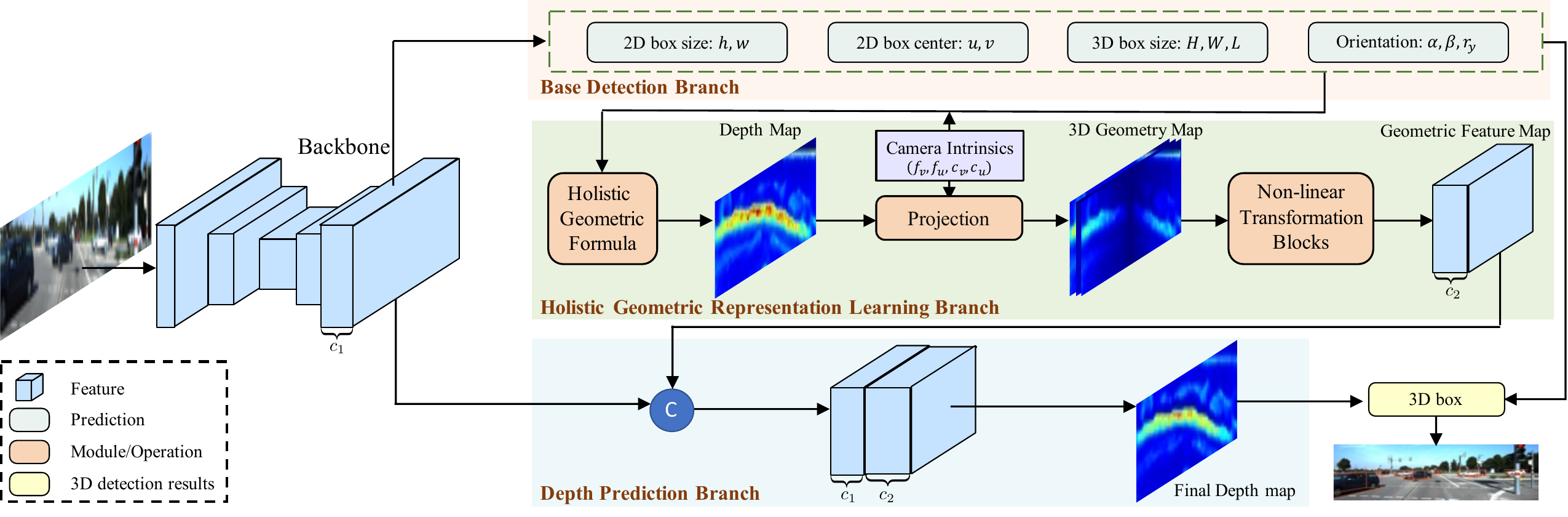}
\end{center}
   \caption{
   \textbf{An overview of our proposed holistic geometric representation learning approach.}
We leverage a network to extract features from the monocular image. 
Then base detection branch is used for generating 2D/3D predictions with depth excluded from image features.
The 2D/3D predictions are utilized by the holistic geometric representation learning branch to generate geometric features via the proposed holistic geometric formula implemented in a network module.
The geometric features are concatenated with the image features from the backbone for depth estimation. 
Based on the depth and other 3D predictions from the base detection branch, the detectors outputs the 3D object detection results. The symbol \textcircled{c} indicates a concatenation operation.
}
\label{fig: framework}
\end{figure*}

\par\noindent\textbf{Geometry-Guided 3D Object Detection.}
There are several recent methods considering utilizing the geometric information for monocular 3D object detection~\cite{monopsr,monogrnet,deep3dbox,decoupled,gs3d}.  
One research direction mainly focuses on using geometry information to improve the detection performance in the inference stage via {post-processing}~\cite{m3drpn, ur3d}. For instance, M3D-RPN~\cite{m3drpn} employs the consistency between the 2D projected and the predicted 2D bounding boxes to optimize orientation parameters in a post-processing process. UR3D~\cite{ur3d} uses estimated key points to post-optimize the predictions of physical sizes and yaw angles by minimizing the objective function. 
Some other works~\cite{monogrnet,deep3dbox,gs3d,decoupled} consider using a simplified perspective projection relationship in the training phase. 
In particular, MonoGRNet~\cite{monogrnet} presents a geometric reasoning method based on instance depth estimation and 2D bounding box projection to obtain more accurate 3D localization.
GS3D~\cite{gs3d} uses average object sizes based on the statistics on the training data to guide the location estimation.
~{
Decoupled-3D~\cite{decoupled} estimates the depth from the projected average height of each vertical edge and the 3D height of the objects. 
RTM3D~\cite{rtm3d} predicts keypoints including eight vertexes and the center of the 3D object in the image plane, and then minimizes the energy function using geometric constraints of perspective projection.
Ivan \etal ~\cite{keypoint} relies on extra CAD models to process labels for keypoint detection and enforces the constrain between 2D keypoints and the CAD models using a consistency loss.
However, 
these methods basically utilize the geometry at the prediction level and ignore several important geometry elements (\eg object poses and positions) in their geometric modeling.}~As discussed before, these works do not consider object pose or dimension. In contrast, our holistic geometric representation jointly models the geometry relationships between the depth and 2D bounding boxes, 3D dimensions, and object poses. 
Besides, most existing works use geometry relationships during post-processing, while our geometric model is implemented as a network module of an end-to-end network to be leveraged for geometry-aware representation learning to directly boost the depth estimation. 
\section{The Proposed Approach}
\subsection{Framework Overview}
An overview of our end-to-end network is shown in Fig.~\ref{fig: framework}. We model an object as a single point following~\cite{centernet, monopair}. 
First, we use deep layer aggregation~\cite{dla} as the backbone to extract features from a monocular image. Second, the features are fed into the base detection branch to separately predict the 2D bounding box, 3D object dimension, and orientation (Sec.~\ref{sec: detection}). 
Third, the holistic geometric representation learning branch models the geometry relationships from these 2D/3D predictions to obtain a holistic geometric formula, which is implemented as a network module for geometry-aware feature learning (Sec.~\ref{sec: geometric formula}). Finally, we utilize the geometric features for depth estimation (Sec.~\ref{sec: geometric formula}), 
which combines with other 3D predictions for obtaining the 3D object detection results.

\subsection{Base Detection Structure}
\label{sec: detection}
Our base network structure for 2D detection, 3D dimension, and orientation prediction is derived from the anchor-free 2D object detection~\cite{centernet, fcos} with six output branches. Each branch takes the backbone features as input and uses 3x3 convolution, ReLU, and 1x1 convolution for prediction.  
In the base detection branch, the heatmap branch is used to locate 2D object center; the 2D/3D offset branch is applied for estimating 2D/3D center in 2D image coordinate system;  the 2D box size and the 3D dimension branch predicts the size of 2D bounding box and the 3D dimension of the 3D object, respectively;
Similar to~\cite{deep3dbox, monopair, centernet}, the orientation branch predicts observation angle $\alpha$ of the object via encoding it into scalars.

\subsection{Holistic Geometric Representation Learning}
\label{sec: geometric formula}

In this section, we introduce the proposed geometric formula via modeling the relationships between the depth and 2D/3D predictions and present how it can be applied to learn holistic geometric representations for depth estimation.

\par\noindent
\textbf{Formulation and notation.}
We adopt the 3D object definition described by the KITTI dataset. 
The coordinate system is constructed in meters with the camera center as the origin of the coordinate.
A 3D bounding box is represented as a 7-tuple
$(W, H, L, x, y, z, r_y)$, where $W, H$ and $L$ are the dimensions of the 3D bounding box, \ie width, height, and length, respectively, and 
$(x, y, z)$
is the bottom center coordinate of the 3D bounding box. 
As shown in Fig.~\ref{fig: beta&theta}, $r_y$ denotes the rotation around the Y-axis in the camera coordinate system, in a range of $[-\pi, \pi]$. 
Moreover, to facilitate the introduction of the proposed geometric formula, we define the 2D bounding box with a 4-tuple  
$(w, h, u, v)$ 
	, where 
$(w, h)$
and
$(u, v)$
represent the size and the center of 2D bounding box, respectively.

\begin{figure}[t]
\begin{center}

\includegraphics[width=1\linewidth]{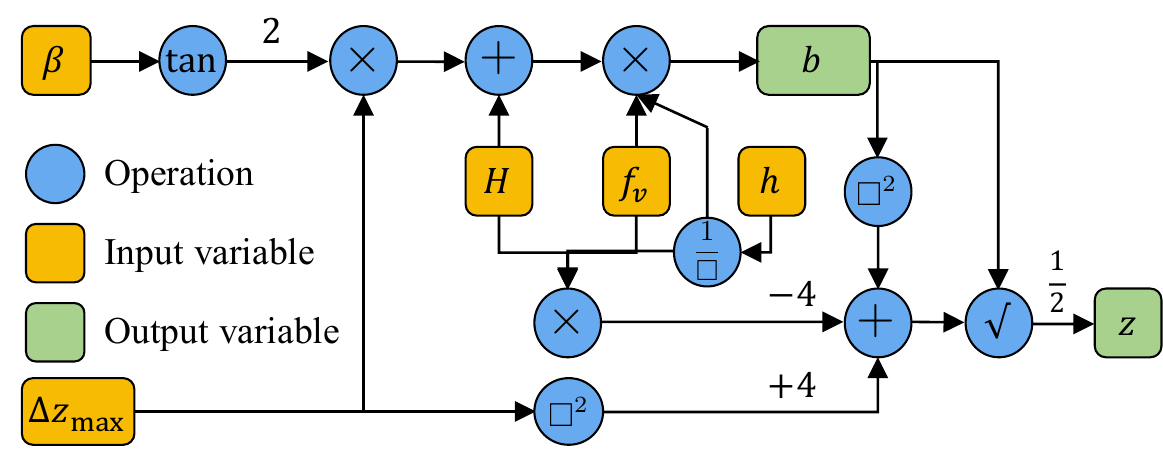}

\end{center}
   \caption{Illustration of the detailed computing flow of the proposed holistic geometric formula as shown in Fig.~\ref{fig: framework}. Particularly, $\Delta z_{\max}$ involves the \emph{3D pose} of objects to represent the maximum difference between the eight corners of the objects in the z-axis, and $\beta$ represents the angle between the bottom center of the object and the horizontal plane.}
\label{fig: compute graph}
\end{figure}

\subsubsection{Holistic Geometric Formula} 
\label{sec: misalignment}
The holistic geometric formula models the object depth according to the geometric relationships between the depth and multiple 2D/3D network predictions.
In particular, the object depth $z$ can be formulated as:
\begin{equation}
\begin{split}
z= \frac{1}{2} b + \frac 1 2 \sqrt{b^2 + 4 (\frac {-H f_v} {h} \Delta z_{max} + \Delta z_{max}^2)},
\end{split}
\label{eq: depth from height}
\end{equation}
where $b = \frac {f_v} {h} (2\tan\beta  * \Delta z_{max} + H)$. $\Delta z_{\max}$ involves the \emph{3D pose} of objects to represent the maximum difference between the eight corners of the objects in the z-axis, and $\beta$ denotes the angle between the bottom center of the object and the horizontal plane.
It can be clearly observed that, the depth $z$ is correlated to the camera intrinsic parameters (\ie $f_v$ and $c_v$), the object position (when deriving $\beta$), 3D dimension (when deriving $\Delta z_{max}$ and $H$), and orientation of the object (when deriving $\Delta z_{max}$). 

In the following, we first elaborate the relationship between the formula in Eq.~\ref{eq: depth from height} and existing works, to better understand the differences between this work and existing ones. Then we show how to derive the holistic geometric formula in Eq.~\ref{eq: depth from height} based on a set of 2D and 3D elements.

\noindent\textbf{Relationship to existing works.}
As we will show later, Eq.~\ref{eq: depth from height} is derived from the perspective projection principle that far-away objects tend to be smaller than the near objects.
If the object is far away from the camera, we can make two assumptions: (i) the object depth will be considerably larger than the object size, and the term $\tan(\beta)$ in Eq.~\ref{eq: depth from height} will be close to zero in this case; (ii) if $\Delta z_{\max} << z$, then we can ignore the influence from $\Delta z_{\max}$ by setting $\Delta z_{\max}=0$. Based on the two assumptions, the holistic formula in Eq.~\ref{eq: depth from height} can be derived as a simplified version (\ie~v2) of the proposed holistic geometric formula:
\begin{equation}
    \setlength{\abovedisplayskip}{5pt}
	z = k * \frac {H} {h},
	\setlength{\belowdisplayskip}{5pt}
\label{eq: s-2}
\end{equation}
where $k$ denotes the factor for the depth scale conversation. The formula in Eq.~\ref{eq: s-2} is widely used in previous works~\cite{gs3d, decoupled}.
The formulation from the previous works~\cite{gs3d, decoupled} in Eq.~\ref{eq: s-2} is clearly different from our formula in Eq.~\ref{eq: depth from height} in two aspects. First,
our formula builds a \emph{non-linear} relationship between the depth $z$ and $H/h$, due to the joint modeling with extra the 3D object pose and the 3D box dimensions, while the formula in Eq.~\ref{eq: s-2} is a linear relationship; Second, the influence of  the factors, \ie~$\Delta z_{\max}$ and $\beta$, on depth calculation is directly ignored by Eq.~\ref{eq: s-2}.

Another possible setting for reducing the computational cost (also investigated in the experimental results) is to consider adopting the first item in Eq.~\ref{eq: depth from height} as another simplified version (\ie~v1) of the proposed holistic geometric formula:
\begin{equation}
    \setlength{\abovedisplayskip}{5pt}
	z= \frac {f_v} {h} (2\tan\beta  * \Delta z_{\max} + H).
	\setlength{\belowdisplayskip}{5pt}
\label{eq: s-1}
\end{equation}
We report detailed comparison between the proposed holistic geometric formulation in Eq.~\ref{eq: depth from height} and the simplified versions (\ie~v1 and v2) in the experiments (see Sec.~\ref{sec: ablation}). In the following, we derive the formulation in Eq.~\ref{eq: depth from height}.

\begin{figure}[t]
\begin{center}

\subfigure[Bird's-eye view]{
\includegraphics[width=0.45\linewidth]{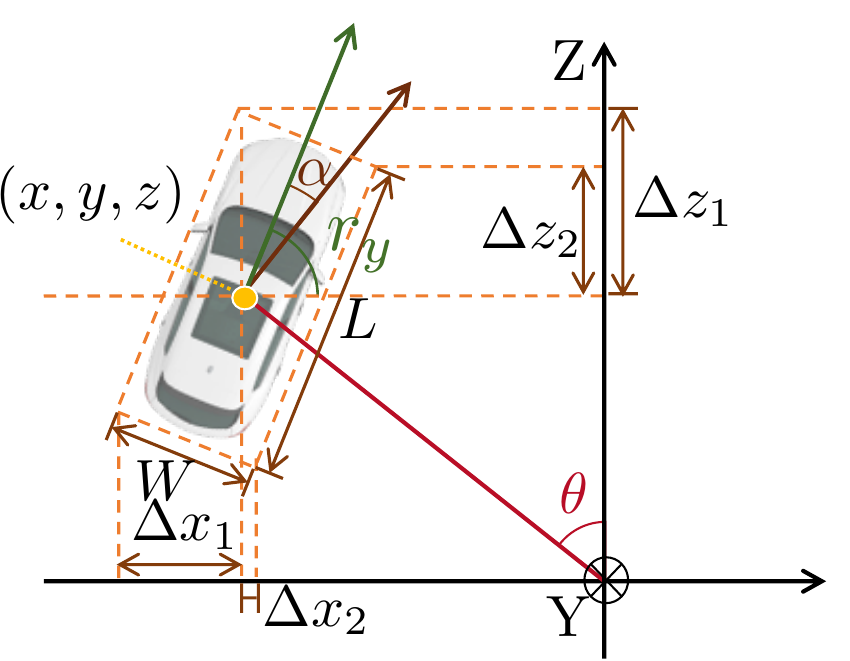}
}\hspace{-0.1 in}
\subfigure[Right side view]{
\includegraphics[width=0.45\linewidth]{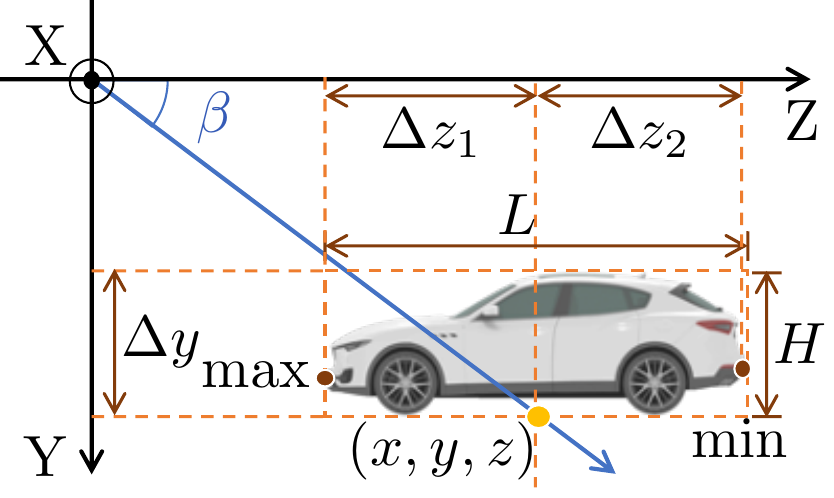}
}
\end{center}

   \caption{Visualization of notations in different object observation angles: (a) $\theta$ in Bird’s Eye View, and (b) $\beta$ in right-side view.}

\label{fig: beta&theta}
\end{figure}

\noindent
\textbf{Geometric relationship of 2D and 3D corners.}
First, we represent an object in the object coordinate system, in which the origin is the bottom center of the object via the translation transformation from the camera coordinate system. 
As shown in Fig.~\ref{fig: beta&theta},
the coordinate of the $c$-th ($c=1,...,8$) corner in the 3D object bounding box, denoted as $\mathbf{P}_{cor}^c$, can be given as follows:
\begin{equation}
\begin{split}
\mathbf{P}_{cor}^c
&=  
\begin{bmatrix}
\pm \Delta x_{i}^c,
+ \Delta y_{i}^c,
\pm \Delta z_{i}^c
\end{bmatrix}^\mathrm{T} \quad \mathrm{s.t.} 
\\[4pt]
\begin{bmatrix}
\Delta x_{i}^c
\\[3pt]
\Delta y_{i}^c
\\[3pt]
\Delta z_{i}^c
\end{bmatrix}
&=
\begin{bmatrix} \frac 1 2 L \cos(r_y)  \pm \frac 1 2 W \sin(r_y)
\\[3pt] 
\frac 1 2 H \pm \frac 1 2 H
\\[3pt]
\frac 1 2 L \sin(r_y)  \mp \frac 1 2 W \cos(r_y)
\end{bmatrix},
\\[3pt]
\end{split}
\label{eq: corner object coordinate}
\end{equation}
where $\Delta x_i^c, \Delta y_i^c$, and $\Delta z_i^c$ represent the coordinate difference between the corner and the center of the object in X, Y, and Z direction, respectively; $i \in \{1, 2\}$ denotes the index of different $\Delta$ values as shown in Fig.~\ref{fig: beta&theta}. 
With the position of the object in the camera coordinate system, we can represent the corner in the same coordinate system as:
\begin{equation}
	\mathbf{P}_{cam}^c 
= \mathbf{P}_{obj} + \mathbf{P}_{cor}^c
= 
\begin{bmatrix} 

x \pm \Delta x_{i}^c
\\
y + \Delta y_{i}^c
\\
z \pm \Delta z_{i}^c
\end{bmatrix},
\label{eq: corner camera coordinate}
\end{equation}
 where $\mathbf{P}_{obj}^c$ and $\mathbf{P}_{cam}^c$ respectively represent the bottom center coordinate and the corner coordinate of the 3D object bounding box in the camera coordinate system; $x$, $y$, and $z$ denote the coordinate value along the X, Y, and Z dimension in the camera plane. $z$ also represents the distance from the bottom center of object to the camera plane, \ie the depth of the object in the camera coordinate system;
Given the intrinsic matrix of the camera provided by the official KITTI dataset, $\mathbf{K}_{inc}$, we can project the corner in the camera coordinate system to the pixel coordinate system as:
\begin{equation}
	\mathbf{P}_{pix}^c
	= [u^c, v^c, 1]^\mathrm{T} = 
	\frac{\mathbf{K}_{inc}\cdot \mathbf{P}_{cam}^c} {z^c},
\label{eq: image coordinate}
\end{equation}
where $\mathbf{P}_{pix}^c$ denotes the projected corner coordinate in the pixel coordinate system; $z^c$ indicates the depth of the $c$-th corner; $u^c$ and $v^c$ respectively denote the horizontal and vertical coordinate of the corner in the pixel coordinate system.

\noindent\textbf{Relationship between 2D height and 3D  corners.}
Given the eight corners of the 3D object box in the pixel plane, 
the height of the projected 2D bounding box $h$ can be estimated from the difference between the vertical coordinate of the uppermost corner (\ie $\max_c\{v^c\}$) and that of the lowermost corner (\ie $\min_c\{v^c\}$) in the pixel coordinate system as:
\begin{equation}
\begin{split}
	h = &\max_c\{v^c\}-\min_c\{v^c\} \\
	 = & \frac {(y + \Delta y_{max} )f_v} {z- \Delta z_{max} } - \frac{(y+ \Delta y_{min})f_v}{z+ \Delta z_{max}} ,
\end{split}
\label{eq: height 2d}
\end{equation}
where $v^c$ is derived from Eq.~\ref{eq: image coordinate}; $\Delta z_{max}=\max_{c}\{\Delta z_i^c\}$ represents the maximum of $\Delta z_i^c$ of the eight corners, analogically for $\Delta y_{max}$;
$f_v$ denotes the focal length in the vertical direction of the pixel plane.

\noindent\textbf{Relationship between depth and other 2D/3D parameters.}
Similar to the definition of the bird's-eye view angle $\theta$ (see Fig.~\ref{fig: beta&theta}a),
we define the angle between the bottom center of the object and the horizontal plane as $\beta$ (see Fig.~\ref{fig: beta&theta}b). Given the projected coordinate $(u_o, v_o)$ of the object bottom center in the pixel plane based on Eq.~\ref{eq: image coordinate}, we can obtain the following geometric relationship:
\begin{equation}
	y = z * \tan(\beta) =z * \frac {v_o - c_v} {f_v},
\label{eq: beta}
\end{equation}
where $c_v$ is the location of the principal point relative to the origin in the pixel plane.
Combining Eq.~\ref{eq: height 2d} and Eq.~\ref{eq: beta}, the depth of the object center, $z$, can be obtained as in Eq.~\ref{eq: depth from height}.

\subsubsection{Learning Enhanced Depth with Holistic Geometric Representations}
Following the proposed geometric formula, we devise and implement a network module for the geometry-guided deep representation learning for accurate depth prediction, as shown in the holistic geometric representation learning branch of Fig.~\ref{fig: framework}. 
The module aims to refine depth estimation through removing input noise introduced by using the 2D/3D geometry-related network predictions
as input.~{Specifically, in both the training and inference stage, the module first produces a calculated one-channel depth map with the proposed geometric formula as described in Eq.~\ref{eq: depth from height}, and a detailed description of the computing flow of the geometric formula is shown in Fig.~\ref{fig: compute graph}. The depth map is then transformed into 3D maps of 3 channels representing a 3D data point $[x, y, z]$ by introducing camera parameters as the initial geometric input. 
Then, the 3D map goes through three non-linear transformation blocks, with each block consisting of convolution, BN, and ReLU layer,
to learn a robust geometric representation map with $C$ channels.
} These geometric representations learned by back-propagation in the end-to-end model are further concatenated with the image representations produced from the backbone network to estimate the enhanced depth.

\begin{table}[t]
\begin{center}
\resizebox{0.99\linewidth}{!}
{
\begin{tabular}{l|ccc|ccc}
\toprule
\multirow{2}{*}{Method}	 &\multicolumn{3}{c|}{3D Detection}	&\multicolumn{3}{c}{BEV}\\
\cline{2-7}
						&Easy &Mod. &Hard	&Easy &Mod. &Hard\\
\hline\hline 	
Baseline    	& 16.42 & 11.84 & 10.06 & 24.47 & 17.17 & 15.40 \\
w/ gt Dim   	& 19.85 & 14.06 & 12.02 & 25.06 & 18.29 & 15.85 \\
w/ gt Depth	 	& \textbf{79.82} & \textbf{70.91} & \textbf{62.41} & \textbf{88.60} & \textbf{82.66} & \textbf{75.41} \\
\bottomrule
\end{tabular}
}
\end{center}
\caption{\textbf{Error analysis.} Similar to the error analysis in~\cite{centernet}, we replace the predicted depth and 3D dimensions with their corresponding ground-truth values. Using the ground-truth depth remarkably improves the AP from 11.84\% to 70.91\% on the moderate, suggesting that the depth is a significantly important factor that affects the accuracy the monocular 3D object detection.
}
\label{tb: error}
\end{table}

\subsection{Misalignment in 2D and 3D bounding Boxes}
There is a misalignment between the 2D projected box and 2D annotation box. 
Generally, due to the perspective projection effect, \ie~further objects appear smaller than nearer objects, the misalignment is more serious for nearby objects, which makes the learning with the proposed formula inaccurate, especially for nearby objects. 
To handle this misalignment, we propose to use the 2D projected box instead of the 2D annotation box as the ground-truth to ensure the correctness of the depth estimation. 
According to Eq.~\ref{eq: corner object coordinate} and ~\ref{eq: corner camera coordinate}, we compute the 3D corner coordinates of the object through the 3D poses and 3D dimensions of the object.
We further obtain their coordinates on the pixel plane through the projection transformation according to Eq.~\ref{eq: image coordinate}. 
We also calculate the difference between vertices in the image plane as the height and width of the 2D projected box.

\subsection{Implementation Details}
\noindent
\textbf{Backbone.}
We adopt a DLA-­34~\cite{dla} network architecture without DCN as our backbone. During training, we set the input resolution of the network as $380 \times 1280$. The spatial size of the feature map from the backbone is 
$\frac{380}{R} \times \frac{1280}{R}$, where $R=4$ represents the down-sampling factor. 

\noindent
\textbf{Optimization loss.}
The optimization objective of our deep detection framework follows a multi-task learning setting, and consist of classification and regression losses for both the 2D and 3D predictions. Specifically, we train the heatmap prediction with the focal loss~\cite{focalloss}.
The branches for offsets and dimensions in both the 2D and 3D detection are trained with $\mathcal{L}$1 losses. The branch for the orientation predictionn is trained with a MultiBin loss following~\cite{monopair, centernet}. 
Based on~\cite{monopair, uncertainty}, we use an $\mathcal{L}$1 loss with heteroscedastic aleatoric uncertainty for the depth estimation.

\begin{table*}[t]
\begin{center}
\resizebox{0.9\linewidth}{!}
{
\begin{tabular}{l |c || ccc| ccc| ccc| c}
\toprule
\multirow{2}{*}{Method}	&\multirow{2}{*}{Extra data}&\multicolumn{3}{c|}{3D Detection}&\multicolumn{3}{c|}{BEV}&\multicolumn{3}{c|}{AOS} &\multirow{2}{*}{Runtime}
\\
\cline{3-11}
			&			&Easy &Mod. &Hard 			&Easy &Mod. &Hard				&Easy &Mod. &Hard			
			\\
\hline\hline
MonoDLE\cite{monodle}	
			&-
    					&17.23  &12.26  &10.29
						&24.79  &18.89  &16.00
						&93.46  &90.23  &80.11
						& - \\
GrooMeD-NMS\cite{groomed-nms}	
			&-
    					&18.10  &12.32  &9.65		
						&26.19  &18.27  &14.05
						&90.05  &79.93  &63.43
						& - \\
DDMP-3D\cite{ddmp-3d}       
            &-
                                &19.71      &12.78      &9.80
                                &28.08      &17.89      &13.44
                                &90.73      &80.20      &61.82
                                &-\\
			\hline\hline
Decoupled-3D\cite{decoupled}	
			&Yes
						&11.08 	&7.02 	&5.63		
						&23.16 	&14.82 	&11.25 
						&87.34 	&67.23 	&53.84
						& - \\
UR3D\cite{ur3d}         &Yes
                        &15.58  &8.61   &6.00
                        & 21.8  &12.51  &9.20
                        &-		&-		&-	
                        &120ms\\
AM3D\cite{am3d}
			&Yes
						&16.50	&10.74	&9.52		
						&{25.03} &{17.32} 	&14.91
						&-		&-		&-			
						& $\sim$400ms\\
PatchNet\cite{patchnet}	
			&Yes
						&15.68	&11.12	&10.17
						&22.97 	&16.86 	&14.97
						&-		&-		&-			
						& $\sim$400ms\\
DA-3Ddet\cite{DA-3Ddet}         & Yes
                                & 16.80     & 11.50     & 8.9
                                &-		&-		&-
                                &-		&-		&-
                                &-\\
D4LCN\cite{d4lcn}	
			&Yes
						&16.65  &11.72	&9.51		
						&22.51 	&16.02 	&12.55
						&90.01  &82.08  &63.98 		
						& - \\
Kinematic3D\cite{kinematic3d}
            &Yes
						& 19.07 & 12.72 & 9.17 
						& 26.69 & 17.52 & 13.10
                        & 58.33 & 45.50 & 34.81
                        & $\sim$120ms \\
CaDDN\cite{CaDDN}
            & Yes
                        &19.17  &13.41  &11.46
                        &27.94  &18.91  &17.19
                        &78.28  &67.31  &59.52
                        & - \\
\hline\hline 
GS3D\cite{gs3d}		
			&No
						&4.47 	&2.90 	&2.47 
						&8.41 	&6.08 	&4.94 		
						&85.79 	&75.63 	&61.85 
						& $\sim$2000ms\\

MonoGRNet\cite{monogrnet}
			&No		&9.61 	&5.74	&4.25		&18.19 	&11.17 	&8.73
						&-		&-		&-			
						& $\sim$60ms \\
MonoDIS\cite{monodis}  	
			&No		&10.37 	&7.94 	&6.40		&17.23 	&13.19 	&11.12
						&-		&-		&-			
						& - \\
M3D-RPN\cite{m3drpn}		
			&No		    &14.76	&9.71 	&7.42		
					 	&21.02  &13.67 	&10.23
						&88.38 	&82.81 	&67.08		
						& 161ms\\
MonoPair\cite{monopair}
			&No		    &13.04	&9.99   &8.65
						&19.28  &14.83  &12.89
						&91.65 	
						&86.11
						&76.45 		
						& 57ms\\
RTM3D\cite{rtm3d}               &No
                                & 14.41     & 10.34     & 8.77
                                & 19.17     & 14.20     & 11.99
                                & \boldblue{91.75}     
                                & \boldblue{86.73}     
                                & \boldblue{77.18}
                                & 55ms\\
MoVi-3D\cite{MoVi-3D}           &No
                                & 15.19     & 10.90      & 9.26
                                & \boldblue{22.76}
                                & \boldblue{17.03}      
                                & \boldblue{14.85}
                                &-		&-		&-
                                & 45ms\\
RAR-Net\cite{liu2020reinforced} &No
                                & \boldblue{16.37}     
                                & \boldblue{11.01}     
                                & \boldblue{9.52}
                                & 22.45     & 15.02     & 12.93
                                & 88.48     & 83.29     & 67.54
                                & - \\						
\hline\hline
Our method	&No		&\textbf{18.85}	&\textbf{13.81}	&\textbf{11.52}							&\textbf{25.86}	&\textbf{18.99}	&\textbf{16.19}	
						&\textbf{94.67}	&\textbf{89.44}	&\textbf{79.27}
						& 50ms\\
Improvement	            &-	
						&+2.48	&+2.80	&+2.00	    
						&+3.10	&+1.96	&+1.34	    
						&+2.92	&+2.71	&+2.09      
						& - \\

\bottomrule
\end{tabular}
}
\end{center}
\caption{
\textbf{State-of-the-art comparison on the KITTI {\textit{test}} set} for the car category in terms of the metric of $\rm {AP}_{40}$. 
Extra data denotes the methods with extra data or external networks used in the training or inference or not. `-' denotes the methods have not been published yet without specific details.
The bold \textbf{black}/\boldblue{blue} color indicates the best/the second best performing method under the same `No' setting. `Improvement' denotes the increasing in performance compared to methods without extra data.
}
\label{tb: test 3d}
\end{table*}

\noindent
\textbf{Training:}
We use a batch size of $32$ and train the overall deep network for 140 epochs on $6$ NVIDIA 1080ti GPUs. To alleviate overfitting, we adopt data augmentation techniques including random scaling, random horizontal flipping, and random cropping for the 2D detection, and random horizontal flipping for the 3D detection, respectively. We use the Adam optimizer with 1e-­5 weight decay to optimize the full training loss as described in~\cite{monopair}. The initial learning rate is 1.25e-­4, which is dropped by multiplying $0.1$ after the $90$-th and the $120$-th epoch. To make train stable, we apply the linear warm-up strategy for learning with the geometric network module in the first 5 epochs.

\noindent
\textbf{Inference:}
We first predict 2D bounding boxes, 3D dimensions, and orientations via a shared backbone and several separate task branches. 
Then, we use the proposed geometric module leveraging the above 2D/3D predictions to predict depth.
Finally, similar to~\cite{centernet}, we use a simple post-processing algorithm through $3\times3$ maxpooling and back-projection to recover 3D bounding boxes from 2D boxes, 3D dimensions, orientations, and the depth.

\section{Experiments}
\noindent\textbf{Setup.} The KITTI dataset~\cite{kitti} provides widely used benchmarks for various visual tasks in the autonomous driving, including 2D Object detection, Average Orientation Similarity (AOS), Bird’s Eye View (BEV), and 3D Object Detection. The official data set contains 7481 training and 7518 test images with 2D and 3D bounding box annotations for cars, pedestrians, and cyclists. We report the average accuracy ($\rm {AP}$) for each task under three different settings: easy, moderate, and hard, as defined in~\cite{kitti}. Moreover, we use 40 recall positions instead of 11 recall positions proposed in the original Pascal VOC benchmark, following~\cite{monodis}. This results in a more fair comparison of the results. Each class uses different IoU standards for further evaluations. We report our results on the official settings of IoU $\geq 0.7$ for cars.

\begin{table*}[t]
\begin{center}
\resizebox{0.93\linewidth}{!}
{
\begin{tabular}{l |ccc |ccc |ccc|ccc}
\toprule
\multirow{2}{*}{Method}	
&\multicolumn{3}{c|}{3D Detection IoU$\geq$0.7}		&\multicolumn{3}{c|}{BEV IoU$\geq$0.7} &\multicolumn{3}{c|}{3D Detection IoU$\geq$0.5}		&\multicolumn{3}{c}{BEV IoU$\geq$0.5}\\
\cline{2-13}
						&Easy &Mod. &Hard 			&Easy &Mod. &Hard
						&Easy &Mod. &Hard 			&Easy &Mod. &Hard\\
\hline\hline 
CenterNet~\cite{centernet}		
						&0.60 	&0.66 	&0.77		&3.46 	&3.31 	&3.21
						&20.00 	&17.50 	&15.57		&34.36 	&27.91 	&24.65\\
MonoGRNet~\cite{monogrnet}			
						&11.90 	&7.56	&5.76		&19.72 	&12.81 	&10.15
						&47.59 	&32.28 	&25.50 		&52.13 	&35.99 	&28.72\\
MonoDIS~\cite{monodis} 				
						&11.06 	&7.60	&6.37		&18.45 	&12.58 	&10.66
						&-		&-		&-			&-		&-		&-	\\
M3D-RPN~\cite{m3drpn}				
						&14.53 	&11.07	&8.65		&20.85 	&15.62 	&11.88
						&48.53 	&35.94 	&28.59		&53.35 	&39.60 	&31.76\\
MoVi-3D~\cite{MoVi-3D}	
                        &14.28  &11.13  &9.68       &22.36  &17.87  &15.73
                        &-		&-		&-			&-		&-		&-	\\
MonoPair~\cite{monopair}				
						&16.28  &12.30  &10.42
						&\boldblue{24.12}
						& 18.17 & 15.76
						&\boldblue{55.38}
						&\boldblue{42.39}
						&\boldblue{37.99}
						&\boldblue{61.06}
						&\boldblue{47.63}
						&\boldblue{41.92} \\
\hline\hline
Baseline			    & \boldblue{16.54}     
                        & \boldblue{13.37}    
                        & \boldblue{11.15}   
                        & 23.62  
                        & \boldblue{19.19}
                        & \boldblue{16.70}
                        & 53.93 	& 40.97 	& 36.67 	& 58.72 	& 45.48 	& 40.02 \\
                         		 	 	 	 
Our method				&\boldblack{18.45}	&\boldblack{14.48}	                               &\boldblack{12.87}
						&\boldblack{27.15}	&\boldblack{21.17}	&\boldblack{18.35}	
						&\boldblack{56.59}	&\boldblack{43.70}	&\boldblack{39.37}	
						&\boldblack{61.96}	&\boldblack{47.84}	&\boldblack{43.10}\\
\bottomrule
\end{tabular}
}
\end{center}
\caption{
\textbf{Monocular 3D object detection results on the KITTI {\textit{val}} set} for the car category with the evaluation metric of $\rm {AP}_{40}$. The results of the previous works are from~\cite{monopair}. Our approach significantly outperforms the previous state-of-the-arts on almost all the different evaluation protocols and settings. The bold \textbf{black}/\boldblue{blue} color indicates the best/the second best performing method.}
\label{tb: val 3d}
\end{table*}

\subsection{Overall Performance Comparison and Analysis}
Table~\ref{tb: test 3d} and~\ref{tb: val 3d} show the overall performance of the proposed approach on the KITTI 3D \textit{test} and \textit{val} sets for cars from the official online leaderboard as of Mar. 12th, 2021. 
Existing state-of-the-art monocular 3D object detectors, including methods using extra data and only using monocular image are listed in the tables. 
There are some methods~\cite{monopair, m3drpn, monogrnet} results on the KITTI \textit{val} quoted from~\cite{monopair}.

\begin{table}[]
\begin{center}
\resizebox{1\linewidth}{!}
{
\begin{tabular}{l|ccc|ccc}
\toprule
\multirow{2}{*}{description}  & \multicolumn{3}{c}{3D Detection} & \multicolumn{3}{c}{BEV} \\
\cline{2-7}
    & Easy  & Mod. & Hard     & Easy   & Mod.    & Hard  \\
\hline\hline
Original baseline             & 12.78 	& 9.83 	& 8.27 	& 18.32 	& 14.18 	& 12.11         \\
+ Uncertainty                      & 15.40     & 11.10     & 9.58     & 22.33  & 16.53  & 14.18 \\
+ Center3d                                   & 16.22     & 12.88     & 10.94    & 22.61  & 17.89  & 16.17 \\
+ Projected box                                & 16.54     & 13.37     & 11.15    & 23.62  & 19.19  & 16.70 \\
\hline
Enhanced baseline                  & \textbf{16.54}     & \textbf{13.37}     & \textbf{11.15}    & \textbf{23.62}  & \textbf{19.19}  & \textbf{16.70} \\
\bottomrule
\end{tabular}
}
\end{center}
\caption{\textbf{~{Results of the enhanced baseline }on KITTI \emph{val} set} for the car category with the evalution metric of $\rm {AP}_{40}$. Each row adds an extra component to the above row.}
\label{tb: baseline}
\end{table}
\noindent
\textbf{Build a simple yet strong baseline for monocular 3D object detection.} 
We report the enhanced baseline results of 3D monocular object detection in Table~\ref{tb: baseline}.~{
Overall, the baseline significantly increases the $\rm {AP}_{40}$ performance upon the original one by 3.76\%, 3.54\%, 2.88\% on easy, moderate and hard difficulty levels, respectively. 
This is achieved by introducing three methods to the original baseline.
\emph{First,} we adopt the $\mathcal{L}$1 loss with the aleatoric uncertainty in~\cite{monopair, uncertainty}, which makes training stage more robust to noise input.
\emph{Second,} we use the projected 3D center as the ground-truth for 2D heatmap prediction similar to SMOKE~\cite{smoke}.
\emph{Third,} we address the misalignment between 2D ground-truth bounding boxes and the 2D projection bounding boxes by using 2D projected box as the ground-truth.
This guarantees the consistency between 2D and 3D boxes from the projection relationships in the proposed geometric formula, and ensure the robust learning with the formula.
The enhanced baseline achieves 16.54\%, 13.37\%, 11.15\% on easy, moderate and hard difficulty levels, respectively.
}

\noindent
\textbf{Comparison with monocular image based methods.}
Our approach achieves a notable improvement over the state-of-the-art monocular image-based detectors~\cite{monodis,monogrnet,m3drpn,monopair} on both the \textit{val} and \textit{test} sets.
As shown in Table~\ref{tb: test 3d}, the performance of our approach on the KITTI \textit{test} set,
for the detection on the car category, an indispensable part of the 3D object detection task for the autonomous driving scenario, 
our method achieves 18.85\% ($2.48\%$ improvement) on the easy, 13.81\% ($ 2.80\%$ improvement) on the moderate, and 11.52\% ($ 2.00\%$ improvement) on the hard compared with the previous state-of-the-art image-only method. Besides, compared with unpublished~\cite{monodle, groomed-nms} our method still increases the $\rm {AP}_{40}$ by 1.49 \% on moderate.
For the Bird's Eye View (BEV) on the car class, our method also achieves the best performance, increasing the $\rm {AP}_{40}$ over the second best method by 3.10\%, 1.96\%, 1.34\% on the easy, moderate, and hard level, respectively. 
For the KITTI \textit{val} set, our method also establishes new state-of-the-art performance on both the 3D object detection and the BEV.
Table~\ref{tb: test 3d} and~\ref{tb: val 3d}  shows considerable improvement over the state-of-the-art monocular detection methods with the great robustness, benefiting from the introduction of the proposed geometric formula for learning geometry-aware representations to advance the depth estimation.

\noindent
\textbf{Comparison with methods using extra data or networks.}
The prior methods~\cite{decoupled, am3d, patchnet, d4lcn, CaDDN}
achieve impressive performance on the KITTI \textit{test} set by introducing extra data or external networks. 
Although our method utilizes none of these kinds of information, as shown in Table~\ref{tb: test 3d}, it can still outperform these comparison methods in terms of the $\rm {AP}_{40}$ metric by 0.40\% on the moderate level. 
These significant improvements demonstrate the superior performance of our method with the proposed geometry-guided depth learning for the monocular 3D object detection.

\noindent
\textbf{Latency.}
~{We test our model on Nvidia GTX~1080~Ti, Pytorch~1.1, CUDA~9.0.
As shown in Table~\ref{tb: test 3d}, the proposed method achieves 20 fps and runs similar to other real-time state-of-the-arts~\cite{rtm3d, MoVi-3D}. This clearly demonstrates the efficiency of our method when compared with other competitive methods under the similar experimental environment.
}

\begin{figure*}[t]

\begin{center}
\resizebox{1\linewidth}{!}
{
\subfigure{
\includegraphics[width=0.3\linewidth]{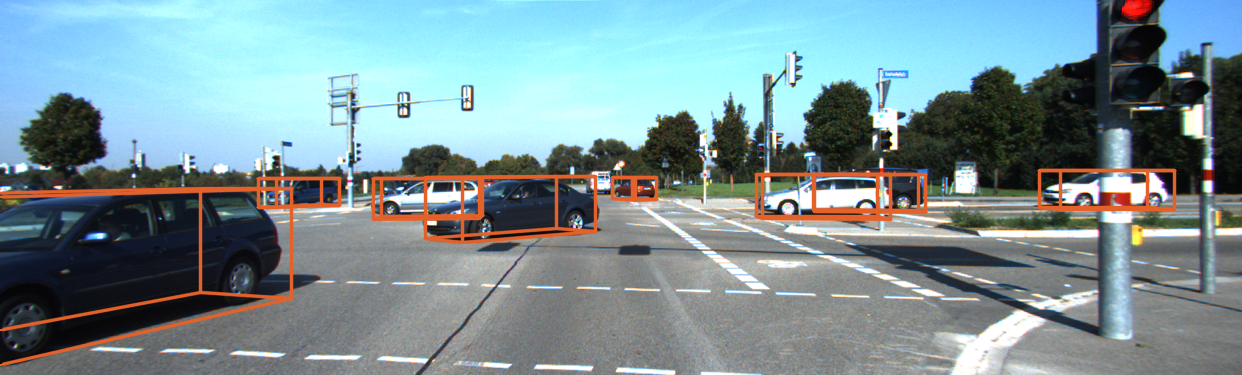}
}\hspace{-0.05 in}
\subfigure{
\includegraphics[width=0.3\linewidth]{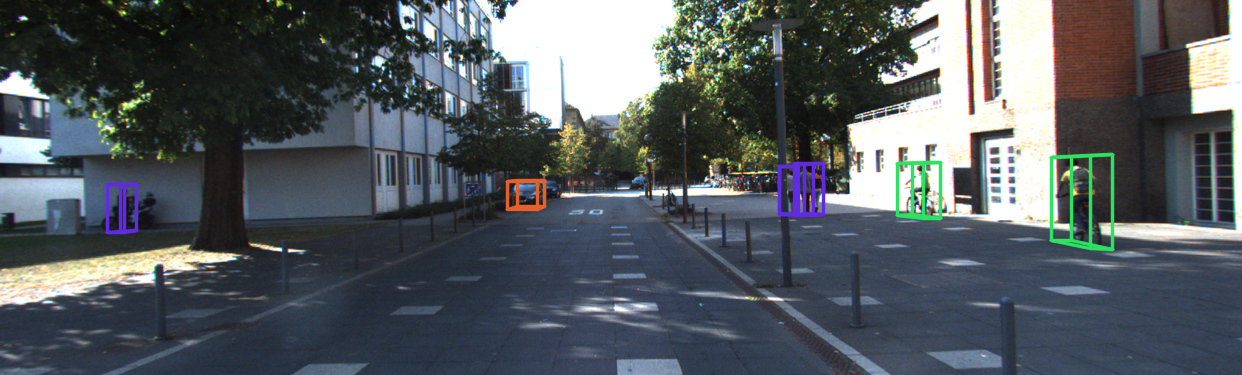}
}\hspace{-0.05 in}
\subfigure{
\includegraphics[width=0.3\linewidth]{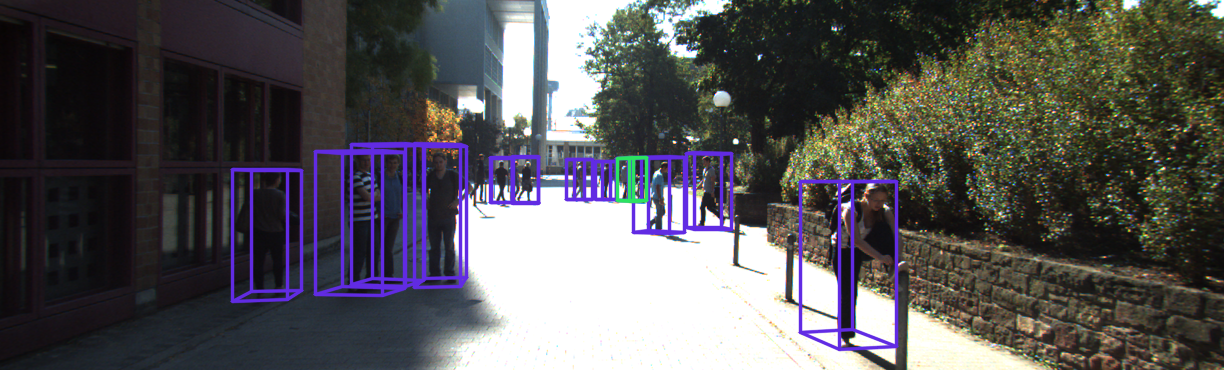}
}
}
\end{center}
   \caption{Qualitative results of our method for multi-class 3D object detection. We use orange box for cars, purple box for pedestrians, and green box for cyclists. All illustrated images are from the KITTI \emph{test} set. Zoom in the image for more details.}
\label{fig: qualitative}
\end{figure*}
\begin{figure*}[t]
\begin{center}
\resizebox{1\linewidth}{!}
{
\subfigure{
\includegraphics[width=0.3\linewidth, trim=0 0 0 50, clip]{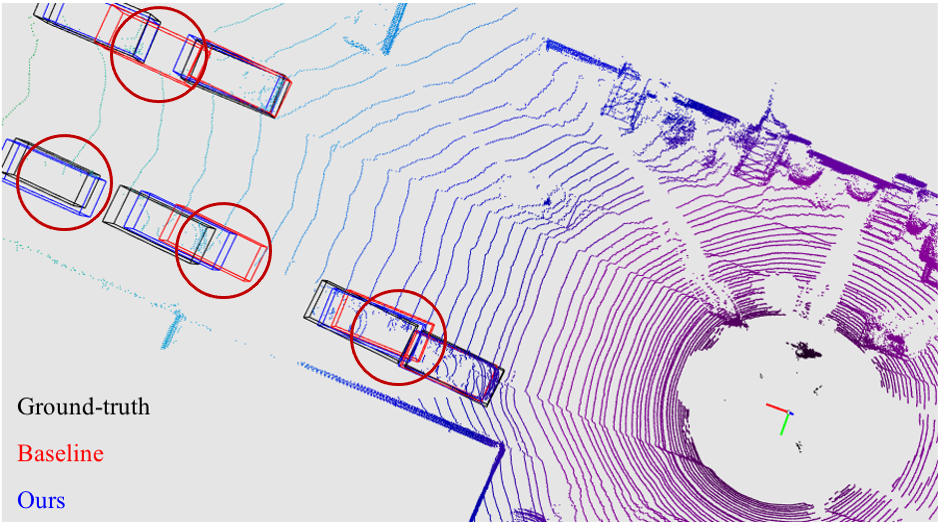}
}
\subfigure{
\includegraphics[width=0.3\linewidth, trim=0 0 0 50, clip]{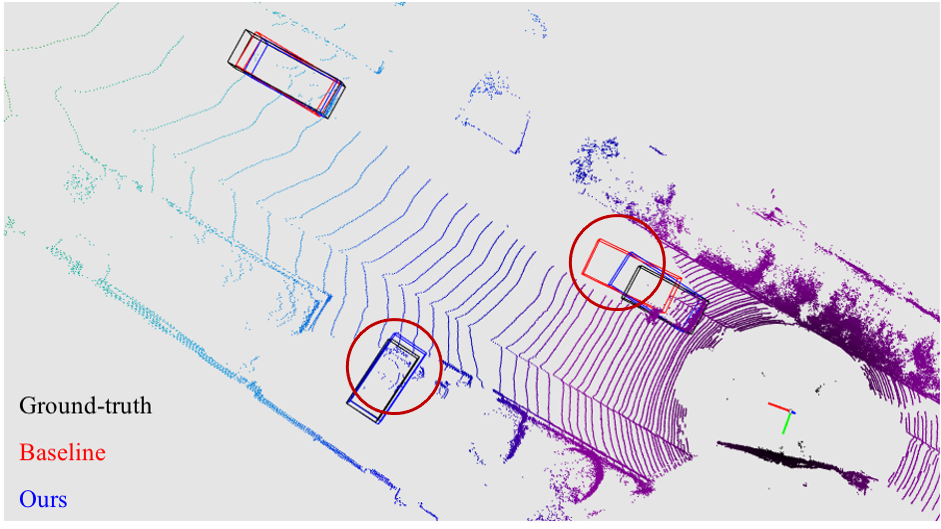}
}
\subfigure{
\includegraphics[width=0.3\linewidth, trim=0 0 0 50, clip]{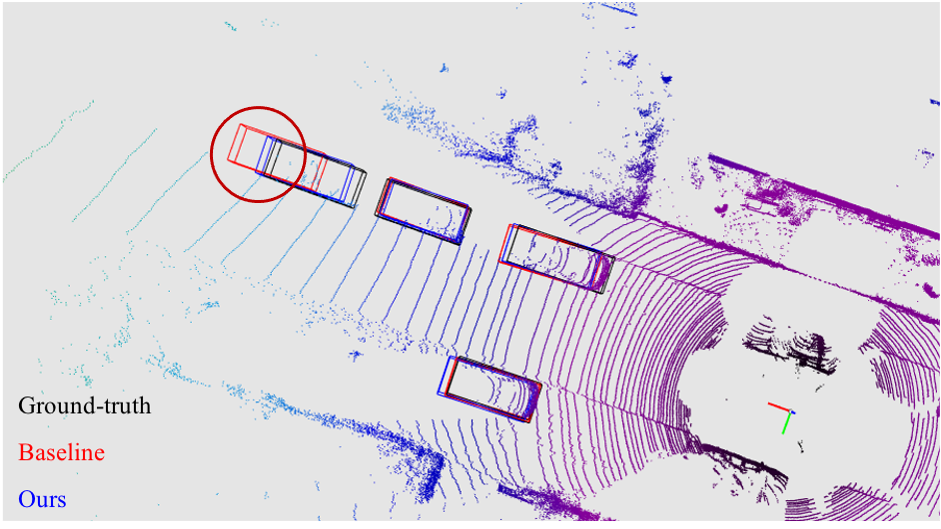}
}
\vspace{-0.1 in}
}
\end{center}
   \caption{{Qualitative} results of our method for Bird’s-Eye-View. We use \textcolor[RGB]{0, 0, 0}{black} box for ground-truth, {red} box for baseline results, and {blue} box for our  results. All the illustrated images are from the KITTI \emph{val} set. Zoom in on the circles for more detailed comparison.}
\label{fig: bev}
\end{figure*}

\begin{table}[t]
\begin{center}
\resizebox{0.9\linewidth}{!}
{
\begin{tabular}{l|ccc|ccc}
\toprule
\multirow{2}{*}{Method} &\multicolumn{3}{c|}{3D Detection} &\multicolumn{3}{c}{BEV}	 \\
\cline{2-7}
					&Easy &Mod. &Hard	&Easy &Mod. &Hard	\\
\hline\hline
Baseline & 16.54 & 13.37     & 11.15    & 23.62  & 19.19  & 16.70 \\
+ 3D-CAT    &15.87 	&11.80 	&10.33 	&21.85 	&16.90 	&14.51  \\
+ Geo-SV1  & 17.25 & 13.38 & 11.29 & 24.33 & 18.57  & 16.06 \\
+ Geo-SV2  & 17.10 & 13.22 & 11.13  & 25.02 & 18.62 & 16.48  \\
Ours (full model) &\boldblack{18.45}	&\boldblack{14.48}	&\boldblack{12.87}
			&\boldblack{27.15}	&\boldblack{21.17}	&\boldblack{18.35} \\
\bottomrule
\end{tabular}
}
\end{center}
\caption{\textbf{Quantitative comparison on different variants of the proposed approach.}~The experiments are conducted on the KITTI \textit{val} set for the car category with the evaluation metric of $\rm {AP}_{40}$, to investigate the effect of the proposed geometric formula and geometry-guided representation learning. `3D-CAT', `Geo-SV1' and `Geo-SV2' represents transformation blocks combined with 3D dimension, simplified geometric formula v1, and v2.}
\label{tb: variants}
\end{table}

\subsection{Ablation Experiments}
\label{sec: ablation}
We conduct extensive ablation studies on the KITTI \textit{val} set, to demonstrate the effectiveness of the proposed approach for geometry-guided depth learning in advancing the monocular 3D object detection. 
For all the evaluation, the $\rm {AP}_{40}$ metric is employed. We mainly investigate from two aspects, including the effect of the proposed geometric formula and module, and the effect of the geometry-guided representation learning for depth estimation.

\par\noindent\textbf{Baseline and variant models.} 
\noindent To conduct an extensive evaluation, we consider the following baseline and variant models: (i) Baseline, which is a base model achieving a strong 3D detection performance with an $\rm {AP}_{40}$ of 11.8\% on the moderate; 
(ii) 3D-CAT., which directly inputs the concatenation of the 3D network predictions to the non-linear transformation blocks while bypassing the depth calculation with geometric formula; (iii) Geo-SV1, which uses our simplified geometric formula v1 as in Eq.~\ref{eq: s-1}; (iv) Geo-SV2, which uses our simplified geometric formula v2 as in Eq.~\ref{eq: s-2}.

\par\noindent\textbf{Effects of the geometric formula and module.} 
A detailed ablation study is shown in Table~\ref{tb: variants}. 
As we can observe, 
ours (full model) achieves a large gain (2.68\% on the moderate) over 3D-CAT, meaning that directly using the network predictions are not effective enough for learning the geometric representations, thus verifying the importance of the proposed geometric formula. By comparing Geo-SV2, Geo-SV1, and ours (full model), all these three with the geometric relationships, the performance gradually improves when more geometry elements are involved in modeling, confirming our motivation of modeling between depth and multiple 2D/3D geometry elements, instead of partial of them, \eg only height considered in most existing works~\cite{gs3d, decoupled}~similar to the Geo-SV2. Finally, Ours (full model) is 1.11\% and 1.98\% improvement on the moderate for the 3D detection and BEV, respectively, which adequately demonstrate the effectiveness of our proposed approach.

\begin{figure}[t]
\begin{center}
\subfigure{
\includegraphics[width=0.48\linewidth]{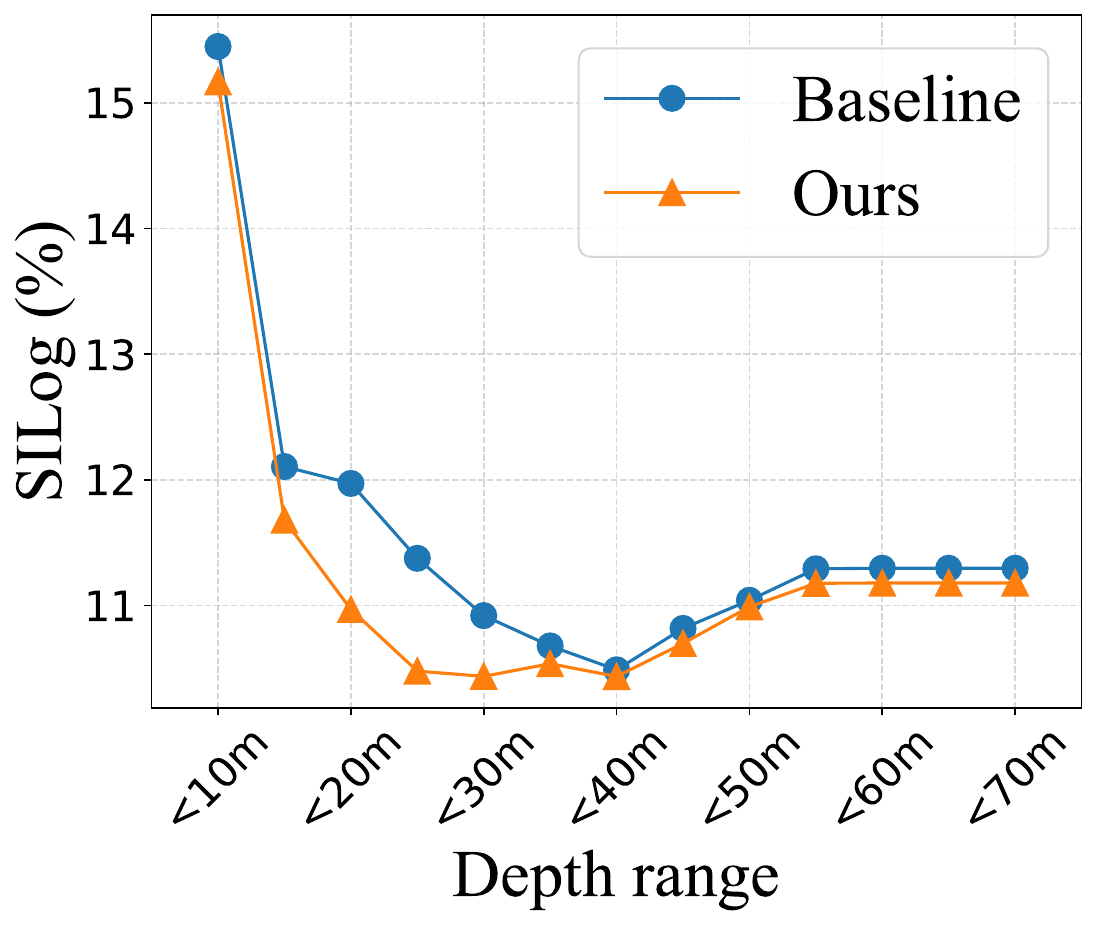}
}\hspace{-0.1 in}
\subfigure{
\includegraphics[width=0.48\linewidth]{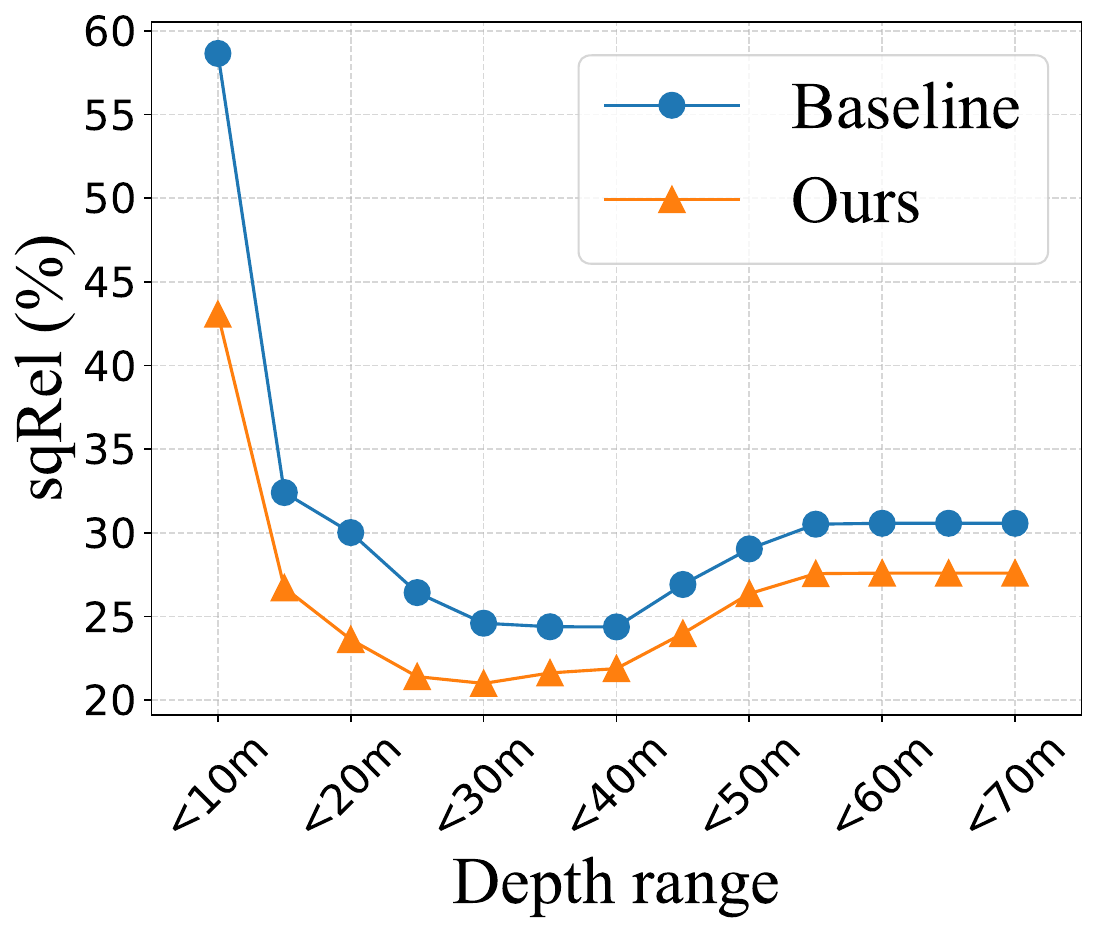}
}
\end{center}
\caption{\textbf{Depth prediction performance} w.r.t. SILog (Scale invariant logarithmic error) and {sqRel} (Relative squared error) metrics on KITTI \emph{val} set for all the car samples. Different depth ranges are considered in the performance evaluation.
}
\label{fig: depth analysis}
\end{figure}
\par\noindent\textbf{Effect of the geometry-guided representation learning for depth estimation.} 
Fig.~\ref{fig: depth analysis} shows a performance comparison between baseline and our approach on the depth estimation. 
Specifically, we evaluate the predicted depth of all car samples in different depth ranges under two primary metrics (\ie SILog and sqRel) widely used in depth estimation field.
On the KITTI train\&val dataset, 87\% of the cars are within 40m, while only 5.0\% of those are 45m away.
Fig.~\ref{fig: depth analysis} shows that our approach outperforms the baseline consistently in all the depth ranges,  especially in the 40m range with most samples, which further validates our idea of using geometry-guided representation learning to boost depth estimation to advance the monocular 3D object detection.

\vspace{-6pt}
\section{Conclusion}
We proposed an effective holistic geometric formula principally modeled from multiple~2D/3D network predictions, to guide the depth estimation and advance the monocular 3D object detection. 
We design and implement this formula as a neural network module to have geometry-aware feature learning with the image representations to boost the learning of the depth. 
Extensive experiments demonstrate the effectiveness of the proposed approach, and results also achieve state-of-the-art performance with a large margin.

\clearpage

\setcounter{table}{0}
\setcounter{section}{0}
\renewcommand\thesection{\Alph{section}} 


In this Supplementary Material, we provide more elaboration on the implementation details, experiment results, and qualitative results.
Specifically, we present the implementation details of the model training in Section~\ref{sec: details},
additional quantitative results, analysis, and limitations in Section~\ref{sec: results},
additional ablation study in Section~\ref{sec: ablation},
and additional qualitative results in Section~\ref{sec: qualitative}.

\section{Additional Implementation Details}
\label{sec: details}

The overall network optimization loss of the proposed approach consists of three parts, \ie~a classification loss $\mathcal{L}_c$, a 2D regression loss $\mathcal{L}_{2D}$, and a 3D regression loss $\mathcal{L}_{3D}$. We present the details of these losses one by one: \textbf{(i)} Regarding to the classification loss, similar to~\cite{cornernet, centernet}, we employ a variant of focal loss $\mathcal{L}_c$ which reduces the penalty for negative locations according to the distance from a positive location as:
\begin{equation}
\mathcal{L}_c =
\begin{cases}
-(1-p)^\alpha \log(p)& \text{if}\ \ y=1\\
-(1-y)^\beta p^\alpha \log(1-p)& \text{otherwise},
\end{cases}
\label{eq: overall loss}
\end{equation}
where $y$ and $p$ represent the ground-truth class probability given by an unnormalized 2D Gaussian and the model's predicted probability for the class, respectively. 
And $\alpha $ and $\beta$ are hyperparameters that control the importance of each sample.
We set $\alpha$ to 2 and $\beta$ to 4 as a default setting in our experiments. \textbf{(ii)} For the 2D regression loss $\mathcal{L}_{2D}$, 
it is defined upon a 6-tuple of ground-truth bounding-box targets and a predicted 6-tuple. Specifically, the 6-tuple consists of two 2D offsets, two 3D offsets, and two 2D box sizes.
2D/3D offsets are used to adjust the 2D/3D center locations before remapping them to the input resolution following~\cite{cornernet, centernet}.
We use an $L1$ loss to optimize each 6-tuple parameters.
\textbf{(iii)} For the 3D regression loss $\mathcal{L}_{3D}$, it consists of an $L1$ loss for regressing the dimension of the 3D bounding box (\ie width, height, and length), and an $L1$ loss with an uncertainty term for regressing the depth. Specifically, we follow~\cite{monopair, uncertainty} and employ the heteroscedastic aleatoric uncertainty in the $L1$ depth estimation loss as:
\begin{align}
	&[\mathbf{d}^*, \sigma^*] 
	            = f^{\theta}(x), \\
	&L(\theta)  =
	\frac {\sqrt{2}} {\sigma} \lVert \mathbf{d} - \mathbf{d}^* \rVert + \log\sigma^*.
\end{align}
Where $\mathbf{d}^*$ and $\mathbf{d}$ represent the predicted depth and the ground-truth depth, respectively. 
$\sigma^*$ is the noisy observation parameter of the model. 
Hence, the overall optimization loss is the sum of the three losses written as:
\begin{equation}
L = \mathcal{L}_c + \lambda_1 \mathcal{L}_{2D} + \lambda_2 \mathcal{L}_{3D}, 
\label{eq: overall loss}
\end{equation}
where $\lambda_1$ and $\lambda_2$ are loss weights controlling the balance between the different losses. We consider $\mathcal{L}_{2D}$ and $\mathcal{L}_{3D}$ equally important and use $\lambda_1$ = $\lambda_2$ = 1 in all experiments.


\section{Additional Results and Analysis}
\label{sec: results}
\subsection{Additional Results for the Pedestrian/Cyclist Category \& Limitations}
\begin{table}[!thb]
\begin{center}
\resizebox{0.9\linewidth}{!}
{
\begin{tabular}{c|c|ccc}
\toprule
\multirow{2}{*}{Cat.} &\multirow{2}{*}{Method}	 &\multicolumn{3}{c}{3D Detection/BEV}\\
\cline{3-5}
					&					&Easy &Mod. &Hard	\\
\hline\hline
\multirow{6}{*}{Ped.}
                    &OFTNet~\cite{oftnet}
                            &0.63/1.28  &0.36/0.81  &0.35/0.51  \\ 
                    &SS3D~\cite{ss3d}
                            &2.31/2.48  &1.78/2.09  &1.48/1.61  \\
	                &M3D-RPN~\cite{m3drpn}			
		 					&4.92/5.65	&3.48/4.05	&2.94/3.29	\\
		 			&MoVi-3D~\cite{MoVi-3D}
		 			        &\boldblue{8.99/10.08} 
		 			        &5.44/6.29  &4.57/5.37  \\
					&MonoPair~\cite{monopair}					
					&\textbf{10.02/10.99	}
					&\textbf{6.68/7.04}
					&\textbf{5.53/6.29}	\\
					&Ours			
					&{8.00/9.54}	
					&\boldblue{5.63/6.77}	
					&\boldblue{4.71/5.83}	\\				
\hline\hline 
\multirow{6}{*}{Cyc.}
                    &OFTNet~\cite{oftnet}
                    &0.14/0.36  &0.06/0.16  &0.07/0.15  \\
                    &SS3D~\cite{ss3d}
                    &2.80/3.45  &1.45/1.89  &1.35/1.44  \\
				 	&M3D-RPN~\cite{m3drpn}			
				 	&0.94/1.25  &0.65/0.81  &0.47/0.78 	\\
				 	&MoVi-3D~\cite{MoVi-3D}
		 			&1.08/1.45  &0.63/0.91  &0.70/0.93  \\
					&MonoPair~\cite{monopair}			
					&\boldblue{3.79/4.76}	
					&\boldblue{2.12/2.87}	
					&\boldblue{1.83/2.42}	\\
					&Ours				
					&\textbf{4.73/5.93}	
					&\textbf{2.93/3.87}	
					&\textbf{2.58/3.42}	\\
\bottomrule
\end{tabular}
}
\end{center}
\vspace{-14pt}
\caption{\textbf{Monocular 3D object detection results on the KITTI \textit{test} set} for the \textbf{Pedestrian and Cyclist} categories with the evaluation metric of $\rm {AP}_{40}$. The IoU threshold is set to 0.5. The bold \textbf{black}/\boldblue{blue} color indicates the best/the second best performing method, respectively.}
\label{tb: pedestrian&cyclist}
\end{table}
\vspace{-10pt}

As mentioned in the main paper, the KITTI~\cite{kitti} official data set contains 7,481 training and 7,518 test images with 2D and 3D bounding box annotations for pedestrian and cyclist categories. 
We report our quantitative results in Table~\ref{tb: pedestrian&cyclist}, using the official settings with IoU $\geq 0.5$ for pedestrians and cyclists on the KITTI \textit{test} set. 
Our method establishes new state-of-the-art performance on all the three detection levels (\ie~easy, medium, and hard) for the cyclist category with only slight drop for the pedestrian category. 
We investigate the slight performance drop in the pedestrian category by comparing 2D detection results between car and pedestrian.
In fact, the advantage of the proposed geometric formula is independent of different classes as 2D images conform with projective camera models, and every object meets the geometric reasoning. 
However, a performance gap between car detection and pedestrian/cyclist detection \emph{commonly exists} in ours and many previous works on the KITTI dataset.
This is mainly due to insufficient training samples of pedestrian and cyclist categories on KITTI, leading to unstable training, sensitivity to hyper-parameters, and inaccurate prediction of 2D/3D information(\eg 2D boxes, orientation, and the 3D dimensions) with high variance.
This imbalance of the category data is however a common issue on the KITTI dataset for the 3D object detection task.
Table~\ref{tb: 2d} shows that the 2D detection results on the moderate level are only
50.48\% and 44.63\% for cyclist and pedestrian respectively, while up to 90.14\% for car on the \textit{test} set.
~Similarly for orientation estimation, the pedestrian~(39.76\%) has less than half of the car~(89.44\%) on the moderate. 
The two factors above introduce more noise into our geometry formula to affect the geometry-guided representation learning.
However, our results for pedestrians and cyclists are highly competitive with other SOTA methods on the KITTI \textit{test} set.
\begin{table}[!thb]
\begin{center}
\resizebox{1.0\linewidth}{!}
{
\begin{tabular}{c|c|ccc}
\toprule
\multirow{2}{*}{Cat.} &\multirow{2}{*}{Method}	 &\multicolumn{3}{c}{2D Detection/AOS}\\
\cline{3-5}
					&					&Easy &Mod. &Hard	\\
\hline\hline
\multirow{3}{*}{Car}
                    &SS3D~\cite{ss3d}
                    &\boldblue{92.72/92.57}
                    &84.92/\boldblue{84.38}  
                    &\boldblue{70.35/69.82}    \\
	                &M3D-RPN~\cite{m3drpn}			
 					&89.04/88.38
 					&\boldblue{85.08}/82.81
 					&69.26/67.08	\\
					&Ours			
					&\boldblack{95.11/94.67}
					&\boldblack{90.14/89.44}
					&\boldblack{80.19/79.27}	\\		
\hline\hline 
\multirow{3}{*}{Ped.}
                    &SS3D~\cite{ss3d}
                    &\boldblack{61.58/53.72}
                    &\boldblack{45.79}/\boldblue{39.60}
                    &\boldblack{41.14}/\boldblue{35.40}    \\
	                &M3D-RPN~\cite{m3drpn}			
		 			&56.64/44.33
		 			&41.46/31.88
		 			&37.31/28.55	\\
					&Ours			
					&\boldblue{58.49/52.87}
					&\boldblue{44.63}/\boldblack{39.76}
					&\boldblue{40.41}/\boldblack{35.83}	\\
\hline\hline 
\multirow{3}{*}{Cyc.}
                    &SS3D~\cite{ss3d}
                    &52.97/42.95
                    &35.48/27.79
                    &31.07/24.26    \\
	                &M3D-RPN~\cite{m3drpn}			
		 			&\boldblue{61.54/48.11}
		 			&\boldblue{41.54/31.09}
		 			&\boldblue{35.23/26.10}	\\
					&Ours			
					&\boldblack{65.42/55.58}
					&\boldblack{50.48/42.05}
					&\boldblack{42.48/35.48}	\\
\bottomrule
\end{tabular}
}
\end{center}
\vspace{-14pt}
\caption{\textbf{Monocular 2D object detection results on the KITTI~\textit{test} set} for the \textbf{All} categories with the evaluation metric of $\rm {AP}_{40}$. The metric $\rm AP_{40}$ is used for detection evaluation and the IoU threshold is set to 0.5. The bold \textbf{black}/\boldblue{blue} color indicates the best/the second best performing method, respectively.}
\label{tb: 2d}
\end{table}
\vspace{-12pt}

\subsection{Further Analysis on Depth Estimation from Geometry Modeling}
We conduct a further depth statistic analysis on the \emph{train+val} set. Table~\ref{tb: analysis_loc_error} shows that for two cars with the same height in both the 2D bounding box and the 3D bounding box, the depth values of their centers may differ by more than $5$ meters due to their distinct \emph{poses} and \emph{locations}.
This confirms the critical importance of considering 3D pose and locations simultaneously in the geometric modeling for depth estimation, which is however not investigated by previous works.

\begin{table}[!thb]
\begin{center}
\resizebox{0.98\linewidth}{!}
{
\begin{tabular}{c|c |c cccc}
\toprule
\multicolumn{1}{l|}{\multirow{2}{*}{$h$}} &\multirow{2}{*}{depth}
	&\multicolumn{5}{c}{The height of 3D bounding boxes}\\
\cline{3-7}
    &   & avg.  & $1.49m$  &$1.50m$  & $1.51m$  & $1.52m$  \\
\hline\hline 
\multirow{3}{*}{30}               
                    & $\max$                    & 39.51 & 40.23 & 40.39 & 42.23 & 39.47 \\
                    & $\min$                    & 37.69 & 36.53 & 36.53 & 37.21 & 37.25 \\
                    & diff.                     & \textbf{1.82}  & \textbf{3.70}  & \textbf{3.86}  & \textbf{5.02}  & \textbf{2.22}  \\
\hline\hline 
\multirow{3}{*}{35}              
                    & $\max$                    & 34.04 & 34.68 & 35.69 & 34.12 & 36.40 \\
                    & $\min$                    & 32.99 & 31.72 & 31.77 & 32.05 & 31.75 \\
                    & diff.                     & \textbf{1.05}  & \textbf{2.96}  & \textbf{3.92}  & \textbf{2.07}  & \textbf{4.65}  \\
\bottomrule
\end{tabular}
}
\end{center}
\vspace{-14pt}
\caption{\textbf{Depth values on the training set (in meter)}. We show the maximum (max) and minimum (min) depth values of the cars with the same height of 2D bounding boxes $h$ and the same height of 3D bounding boxes, and the difference (diff.) between the maximum and minimum depth values.
}
\label{tb: analysis_loc_error}
\vspace{-16pt}
\end{table}

\subsection{Additional Results at Different Distances}
We provide additional results on depth estimation and monocular 3D object detection at different distances.
Table~\ref{tb: depth} shows more depth estimation results on KITTI \emph{val} set via comparing the enhanced baseline and our method.~Specifically, we evaluate the depth estimation by computing Scale Invariant Logarithmic (SILog) error, squared Relative (sqRel) error, absolute Relative (absRel) error, and Root Mean Squared Error of the inverse depth (iRMSE). Our method outperforms the enhanced baseline by large margins on all these evaluation metrics. The depth estimation results clearly demonstrate the effectiveness of our proposed idea of using geometry-guided representation learning to boost depth estimation from monocular images for advancing the monocular 3D object detection.
\begin{table}[!thb]
\begin{center}
\resizebox{1.0\linewidth}{!}
{
\begin{tabular}{c|c|ccccc}
\toprule
Depth Range                  & Num.                   & SILog$\downarrow$ & absRel$\downarrow$ & sqRel$\downarrow$ & iRMSE$\downarrow$ \\
\hline\hline
\multirow{2}{*}{0-10m}  & \multirow{2}{*}{867}   & 16.49 & 8.65   & 67.55 & 16.53 \\
                      &                          & \boldblack{14.35}    & \boldblack{7.75}   
                                                & \boldblack{38.46}     & \boldblack{14.94} \\
\hline
\multirow{2}{*}{0-20m} & \multirow{2}{*}{4236}  & 12.12 & 6.02   & 31.42 & 9.98  \\
                      &                         & \boldblack{10.48} & \boldblack{5.42}
                                                & \boldblack{20.81} & \boldblack{8.77}  \\
\hline
\multirow{2}{*}{0-30m} & \multirow{2}{*}{7379}  & 11.00 & 5.70   & 25.25 & 8.16  \\
                      &                         & \boldblack{10.27} & \boldblack{5.30}
                                                & \boldblack{19.73} & \boldblack{7.52}  \\
\hline
\multirow{2}{*}{0-40m} & \multirow{2}{*}{9797}  & \boldblack{10.49} & 5.65   & 24.68 & 7.23  \\
                      &                         & \boldblack{10.49} & \boldblack{5.36}
                                                & \boldblack{21.71} & \boldblack{7.03}  \\
\bottomrule
\end{tabular}
}
\end{center}
\vspace{-14pt}
\caption{\textbf{Depth prediction results on the KITTI \emph{val} set for all car samples.}~
We show first the baseline and then ours~(\boldblack{bold}) for each row (\ie~each depth range). `Num.' denotes the number of car samples on \emph{val} set, which has in total 11,178 car samples.}
\label{tb: depth}
\vspace{-8pt}
\end{table}

Moreover, we conduct experiments about the 3D monocular object detection improvement at different distances.
Table~\ref{tab: range} reports performance on $\rm {AP}_{40}$ at different object distance ranges following~\cite{categorical}. It is clear that our method consistently outperforms the baseline at different ranges.

\begin{table}[ht]
\small
\centering
\begin{center}
\resizebox{0.95\linewidth}{!}
{
\begin{tabular}{c|ccc|ccc}
\toprule
\multicolumn{1}{c|}{\multirow{2}{*}{Description}}  & \multicolumn{3}{c|}{3D Detection} & \multicolumn{3}{c}{BEV} \\
\cline{2-7}
\multicolumn{1}{c|}{}      & 15m  & 30m & all      & 15m  & 30m & all   \\
\hline\hline
\hline

Baselne     & 18.85 & 15.42 & 11.32 & 26.95 & 21.94 & 16.82 \\
Ours        & \textbf{22.29} & \textbf{17.38} & \textbf{12.87} 
            & \textbf{31.37} & \textbf{24.82} & \textbf{18.35} \\
\bottomrule
\end{tabular}
}
\end{center}
\vspace{-14pt}
\caption{Performance on KITTI \emph{val} at different ranges. }
\vspace{-16pt}
\label{tab: range}
\end{table}

\section{Additional Ablation Study for Uncertainty and Equation}
\label{sec: ablation}
We investigate the effect of uncertainty with our geometric module as requested on the KITTI~\textit{val} set in Table~\ref{tab: uncert}.
It can be seen that the uncertainty is helpful for learning the geometry, but the main improvement is from the proposed principled geometric modeling.
To further validate the effectiveness of Eq.~(6), we compare all predictions followed by pointwise MLP as the reviewer described with our geometric module in Table~\ref{tab: module}. Ours is significantly better than the pointwise MLP.  

\begin{figure*}[!t]
\begin{center}

\subfigure{
\includegraphics[width=0.48\linewidth]{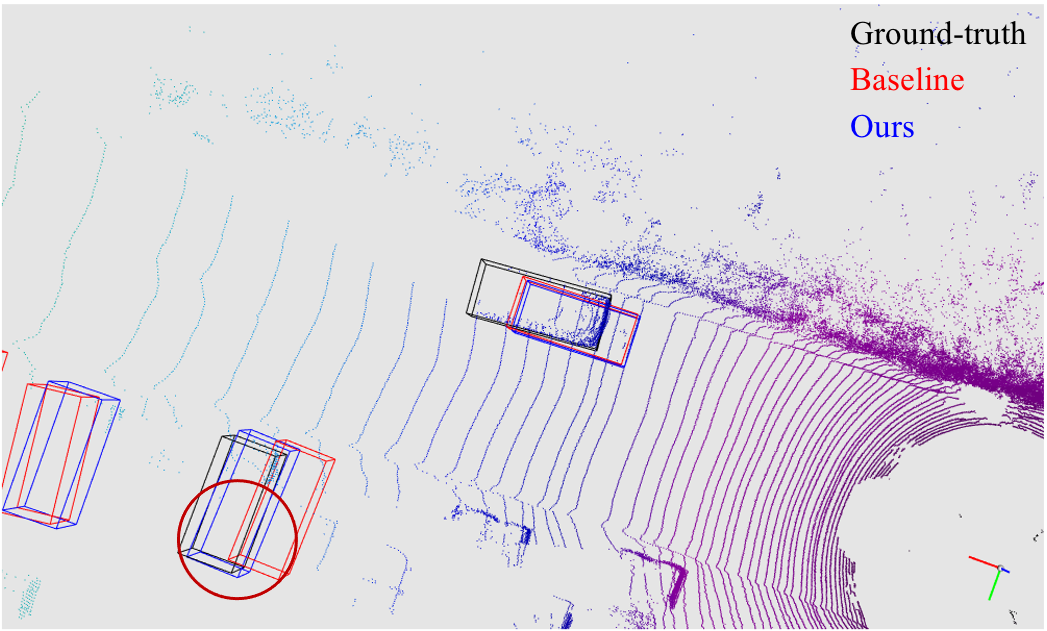}
}\hspace{-0.1 in}
\subfigure{
\includegraphics[width=0.48\linewidth]{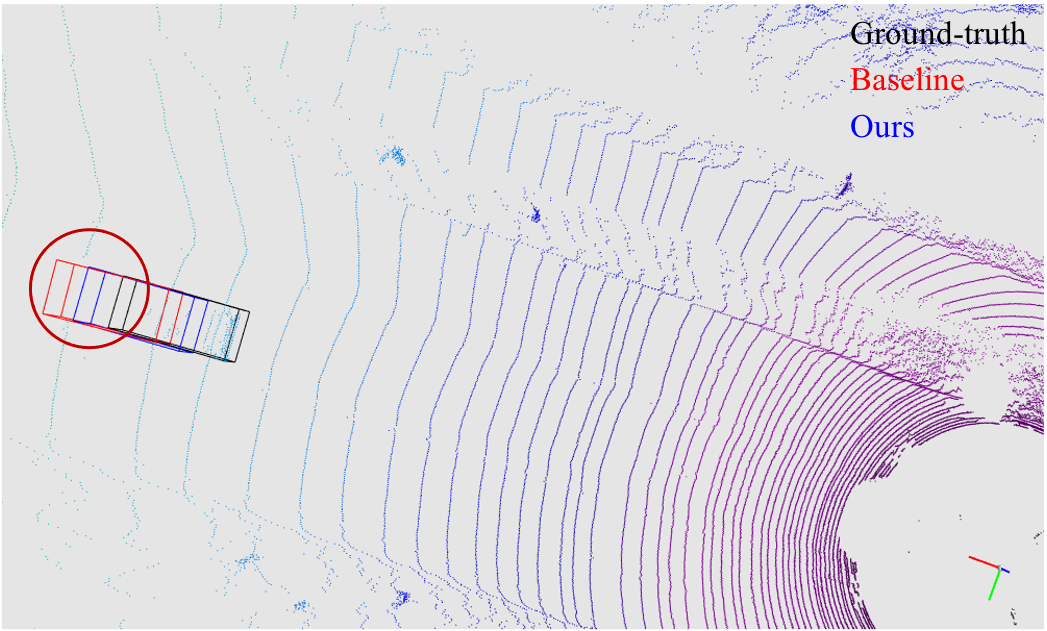}
}
\vspace{-0.1 in}

\subfigure{
\includegraphics[width=0.48\linewidth]{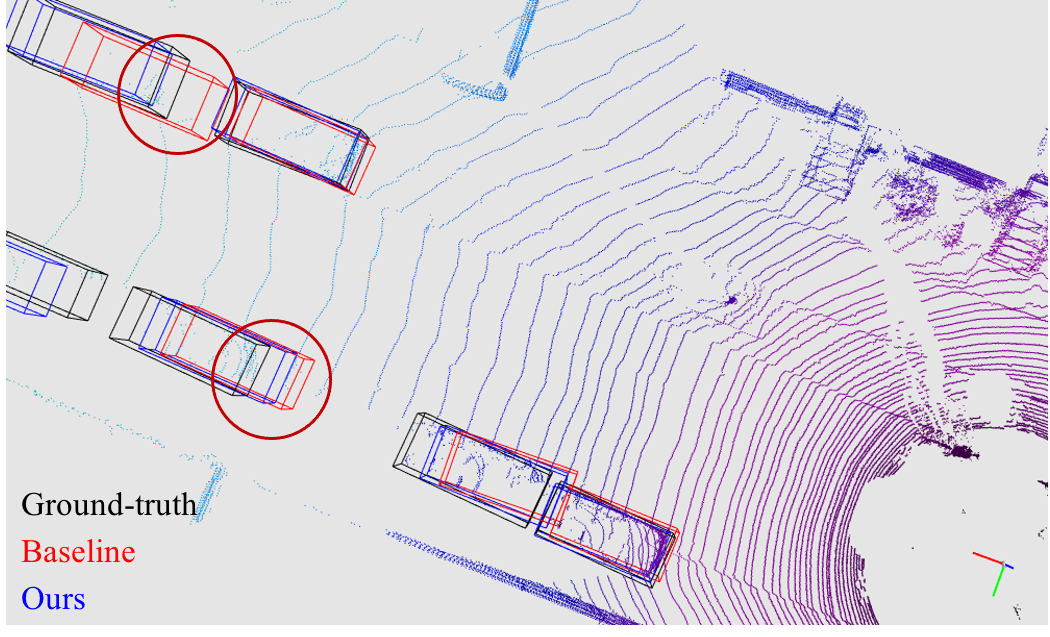}
}\hspace{-0.1 in}
\subfigure{
\includegraphics[width=0.48\linewidth]{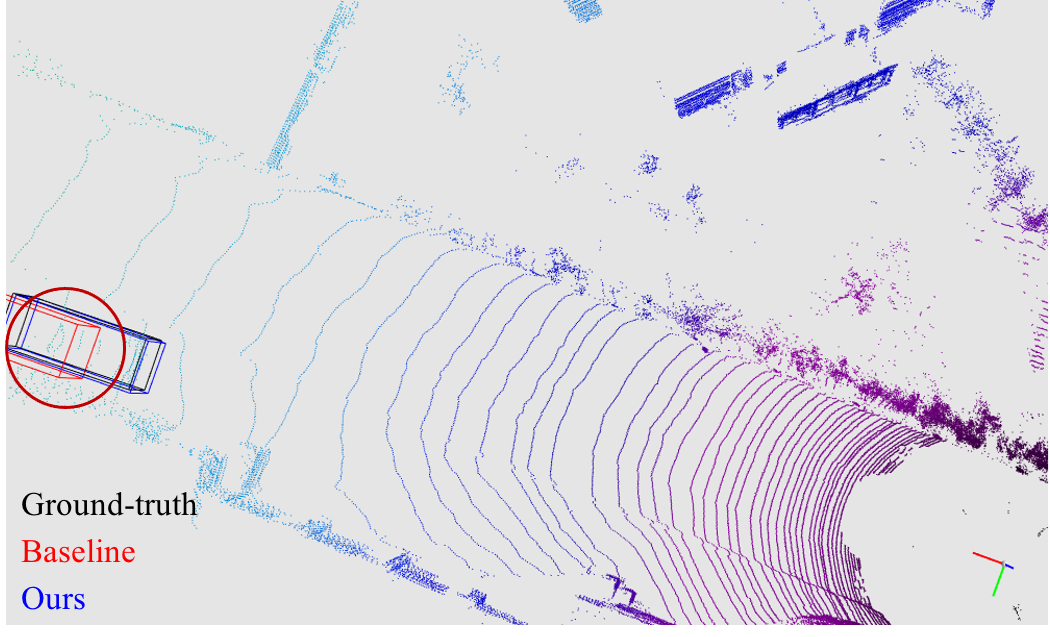}
}\vspace{-0.1 in}
\vspace{-14pt}
\end{center}
  \caption{
Qualitative results of our method for Bird’s-Eye-View. We use \textcolor[RGB]{0, 0, 0}{black} box for ground-truth, {red} box for baseline results, and {blue} box for our  results. All the illustrated images are from the KITTI \emph{val} set. Zoom in on the circles for more detailed comparison.}
\label{fig: bev}
\vspace{-10pt}
\end{figure*}

\section{Additional Qualitative Results}
\label{sec: qualitative}

\begin{table}[!th]
\small
\centering
\begin{center}
\resizebox{0.95\linewidth}{!}
{
\begin{tabular}{c|ccc|ccc}
\toprule
\multicolumn{1}{c|}{\multirow{2}{*}{Description}}  & \multicolumn{3}{c|}{3D Detection} & \multicolumn{3}{c}{BEV} \\
\cline{2-7}
\multicolumn{1}{c|}{}      & Easy  & Mod. & Hard     & Easy   & Mod.    & Hard  \\
\hline\hline
Baseline            & 16.54  	& 13.37  	& 11.15 	
                    & 23.62  	& 19.19  	& 16.70   \\
Pointwise MLP       & 17.09     & 13.12     & 11.05
                    & 23.79     & 18.20     & 16.26   \\
Ours            & \textbf{18.79}     & \textbf{14.53}     & \textbf{12.77}    
                & \textbf{26.48}  & \textbf{20.75}  & \textbf{18.04} \\
\bottomrule
\end{tabular}
}
\end{center}
\vspace{-14pt}
\caption{{~{Results of different modules} on KITTI \emph{val}} with $\rm {AP}_{40}$.}
\label{tab: module}
\vspace{-14pt}
\end{table}

\begin{figure*}[!t]
\begin{center}

\subfigure{
\includegraphics[width=0.49\linewidth]{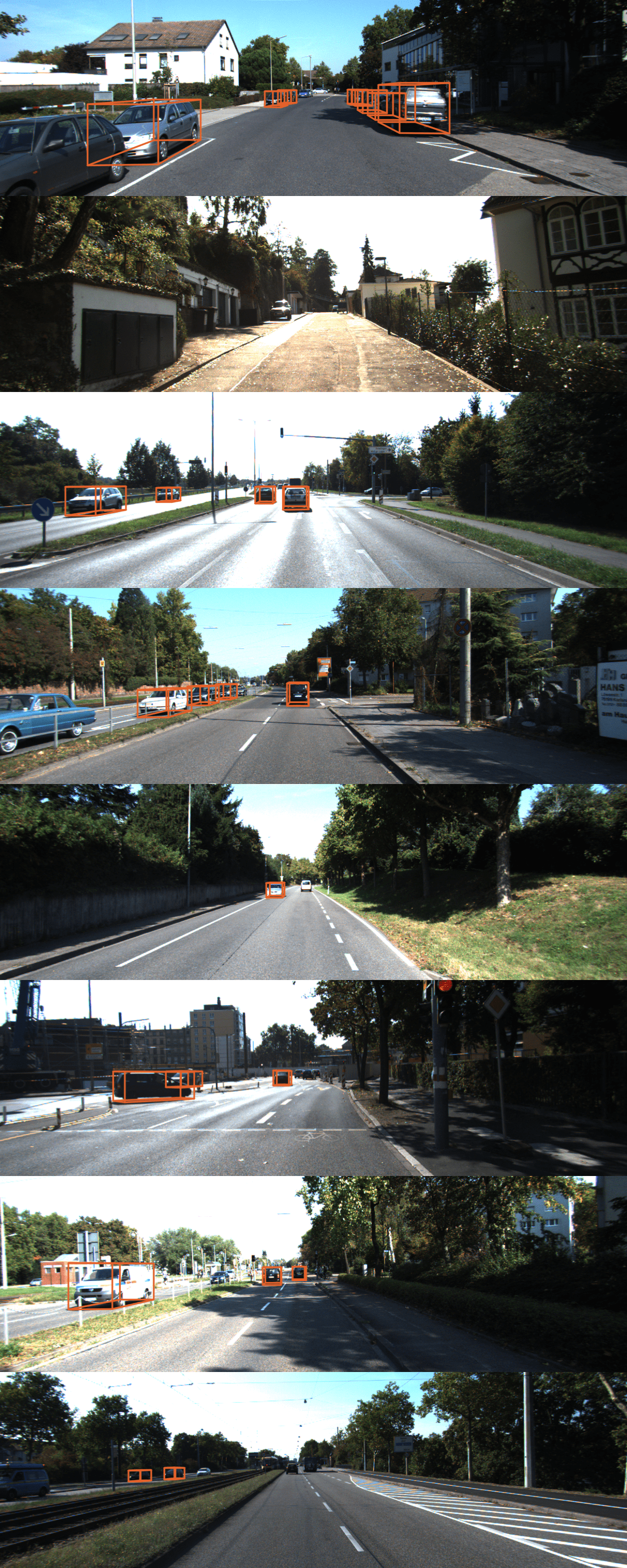}
}\hspace{-0.05 in}
\subfigure{
\includegraphics[width=0.49\linewidth]{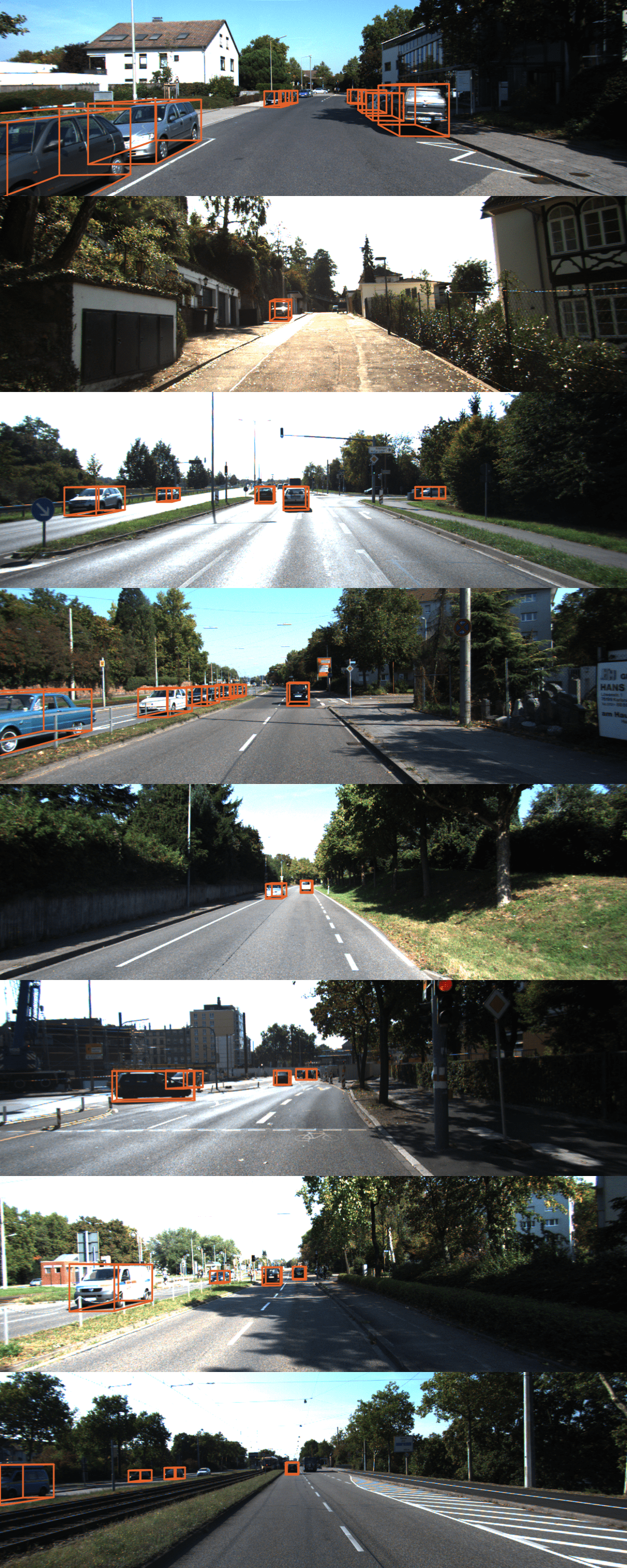}
}\vspace{-0.1 in}

\end{center}
  \caption{
  \textbf{Qualitative Results.}
The predictions on the KITTI \emph{val} set. Results are from the enhanced baseline (left column) and ours (right column).}
\label{fig: val-1}
\end{figure*}

Fig.~\ref{fig: bev} also show the comparison results between the enhanced baseline and the proposed method from the Bird-Eye-View.
Figure~\ref{fig: val-1} also present additional qualitative 3D detection results on the images with a comparison between those two on the KITTI \emph{val} set.

\begin{table}[h]
\small
\centering
\begin{center}
\resizebox{0.95\linewidth}{!}
{
\begin{tabular}{c|c|c|c|c}
\toprule
All Other Enhancements & Uncertainty & Geometric Module & \multicolumn{1}{c|}{3D Detection} & BEV \\
\cline{4-5}
\hline
$\checkmark$ 	&               &              & 11.81 & 17.51 \\
$\checkmark$ 	&               &$\checkmark$  & 14.44 & 19.77 \\  
$\checkmark$    &$\checkmark$   &              & 13.37 & 19.19 \\
              
$\checkmark$ 	&$\checkmark$ &$\checkmark$ & \textbf{14.48} & \textbf{21.17}  \\
\bottomrule
\end{tabular}
}
\end{center}
\vspace{-14pt}
\caption{Ablation study on KITTI \emph{val} set for uncertainty and geometric modeling on the moderate setting of cars.}
\vspace{-14pt}
\label{tab: uncert}
\end{table}

We could observe from the figures that the proposed geometry-guided learning approach can achieve significantly better 3D detection and localization performance than the enhanced baseline.

\begin{figure*}[!t]
\begin{center}
\subfigure{
\includegraphics[width=0.49\linewidth]{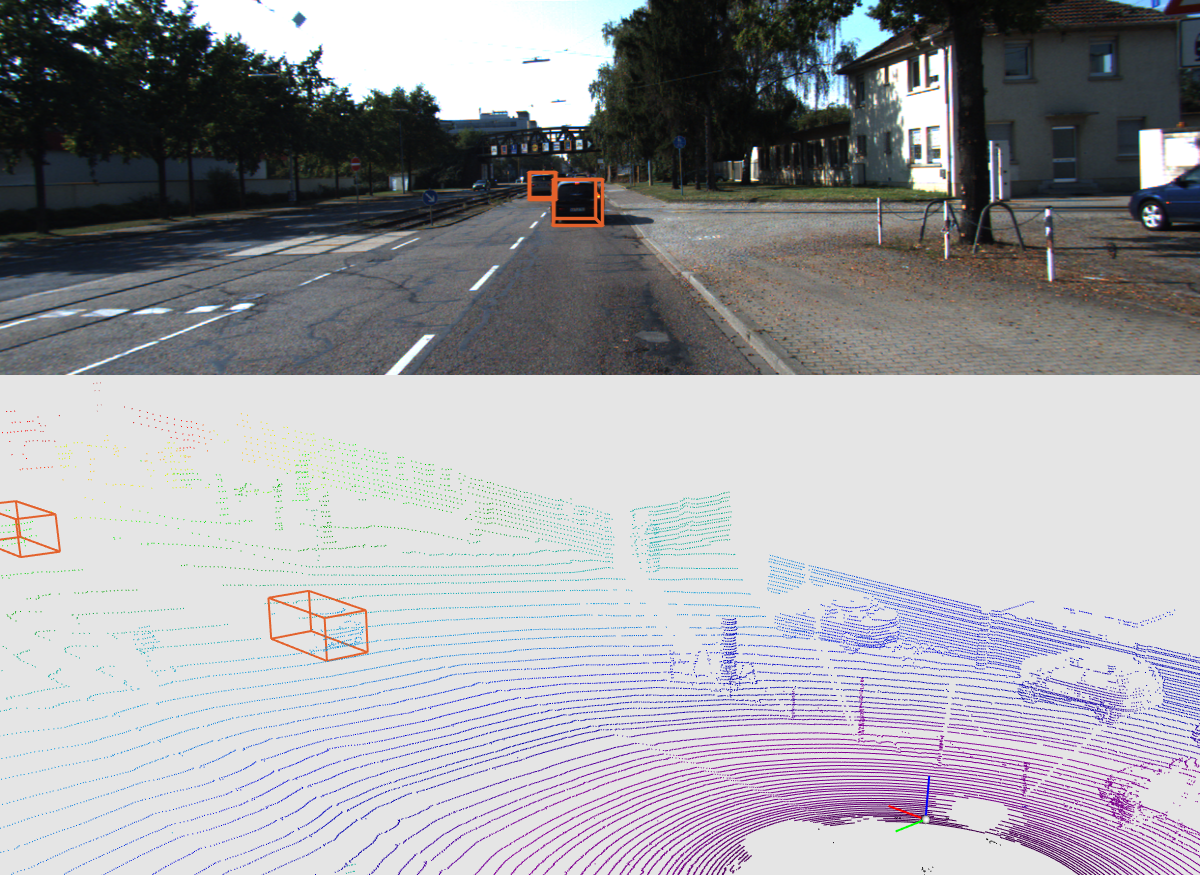}
}\hspace{-0.1 in}
\subfigure{
\includegraphics[width=0.49\linewidth]{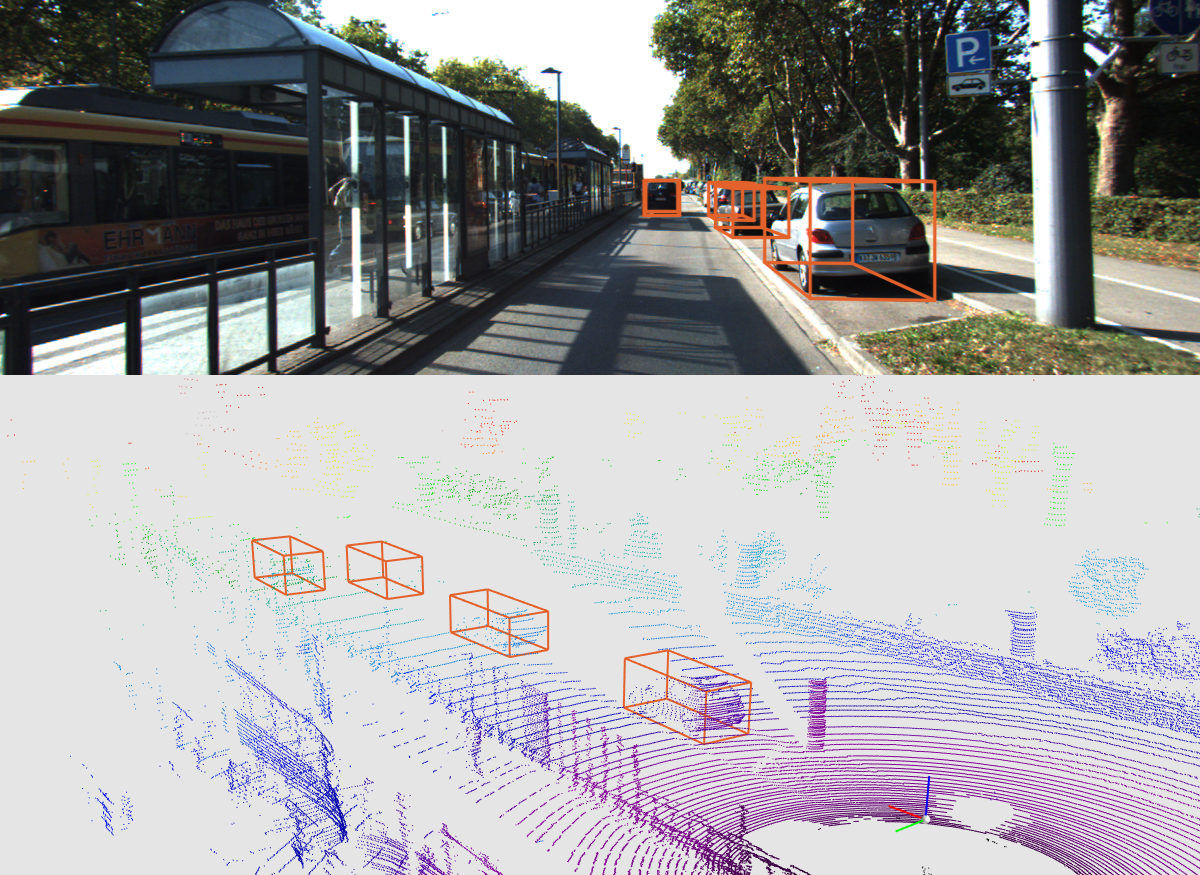}
}

\subfigure{
\includegraphics[width=0.49\linewidth]{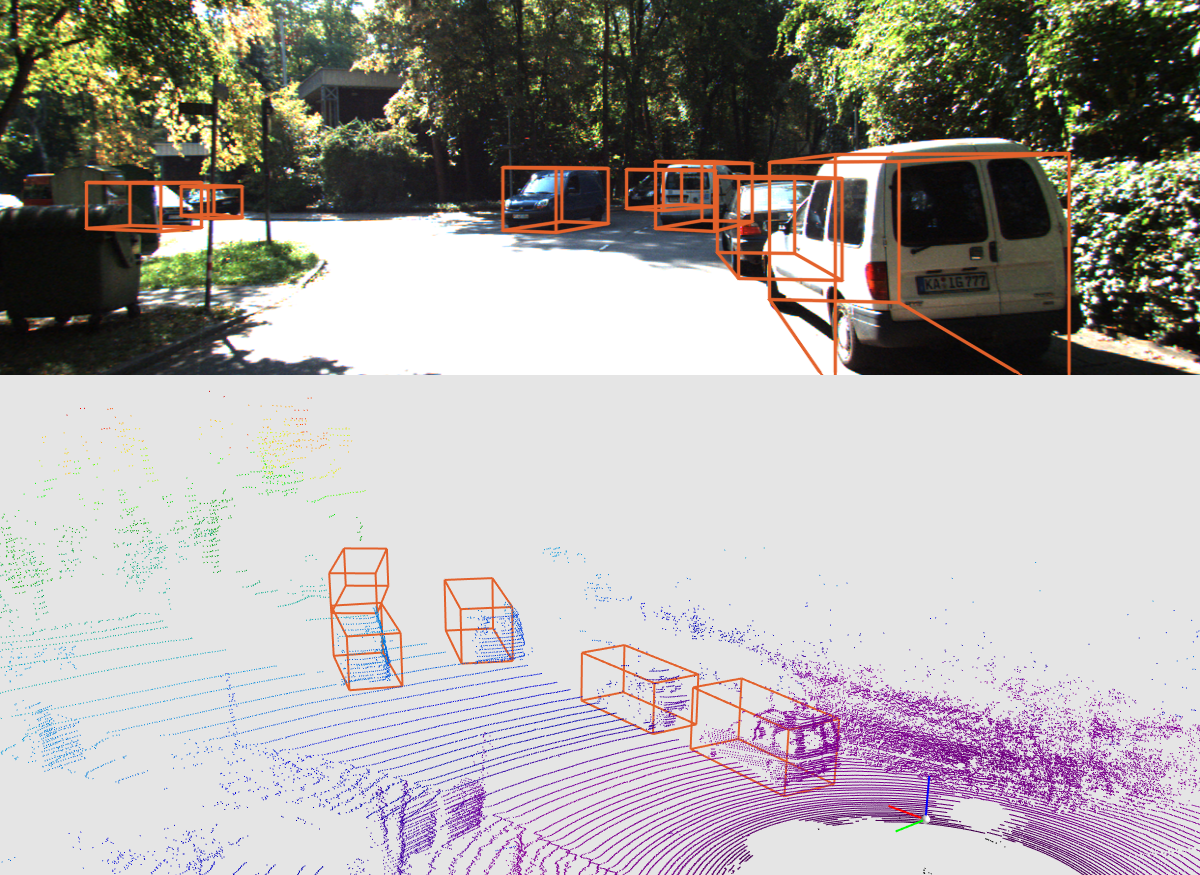}
}\hspace{-0.1 in}
\subfigure{
\includegraphics[width=0.49\linewidth]{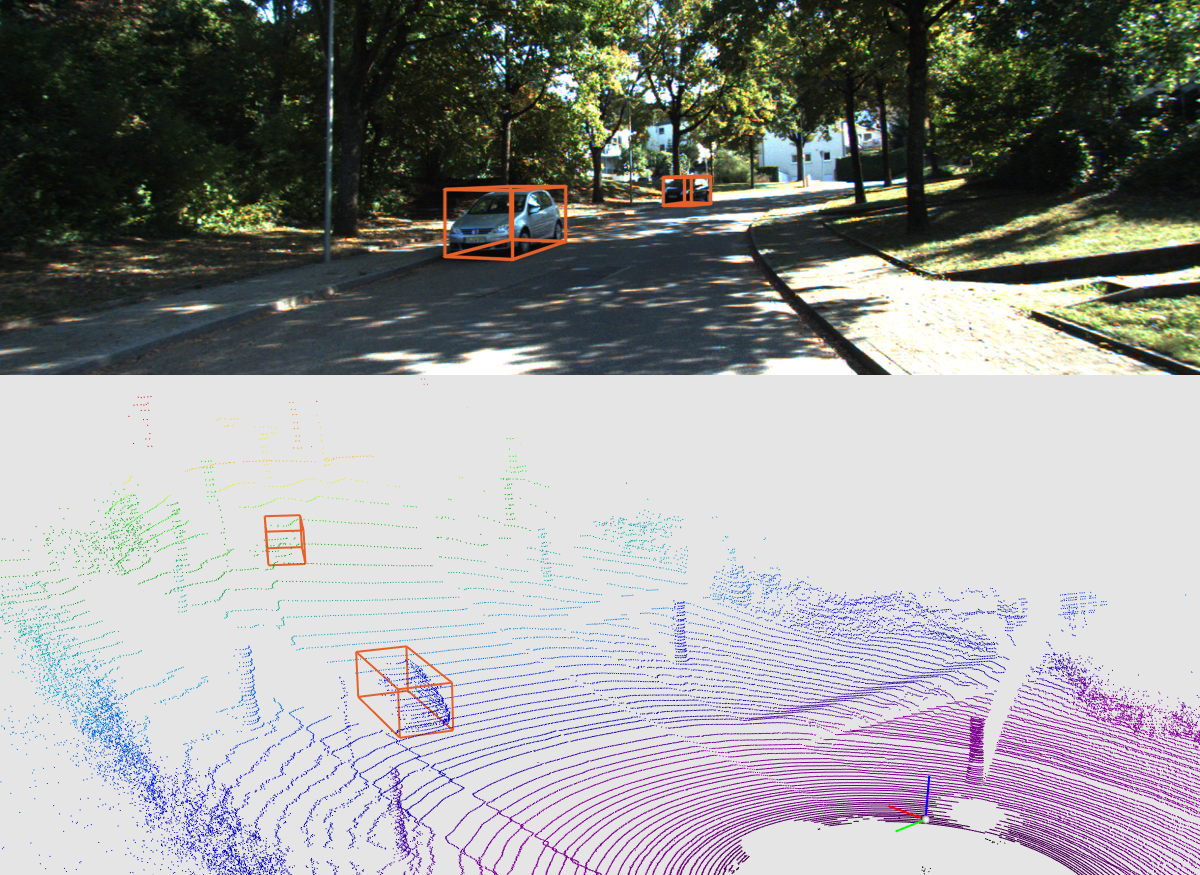}
}

\subfigure{
\includegraphics[width=0.49\linewidth]{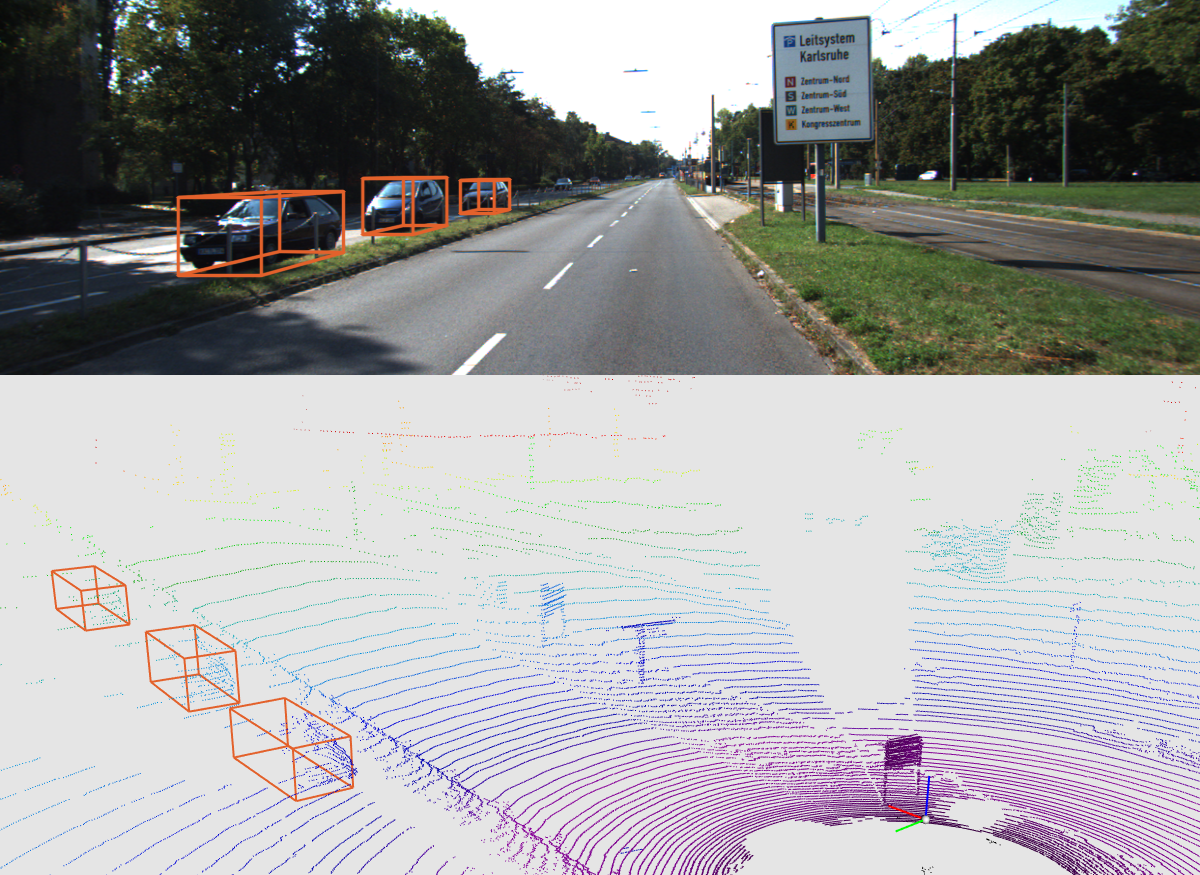}
}\hspace{-0.1 in}
\subfigure{
\includegraphics[width=0.49\linewidth]{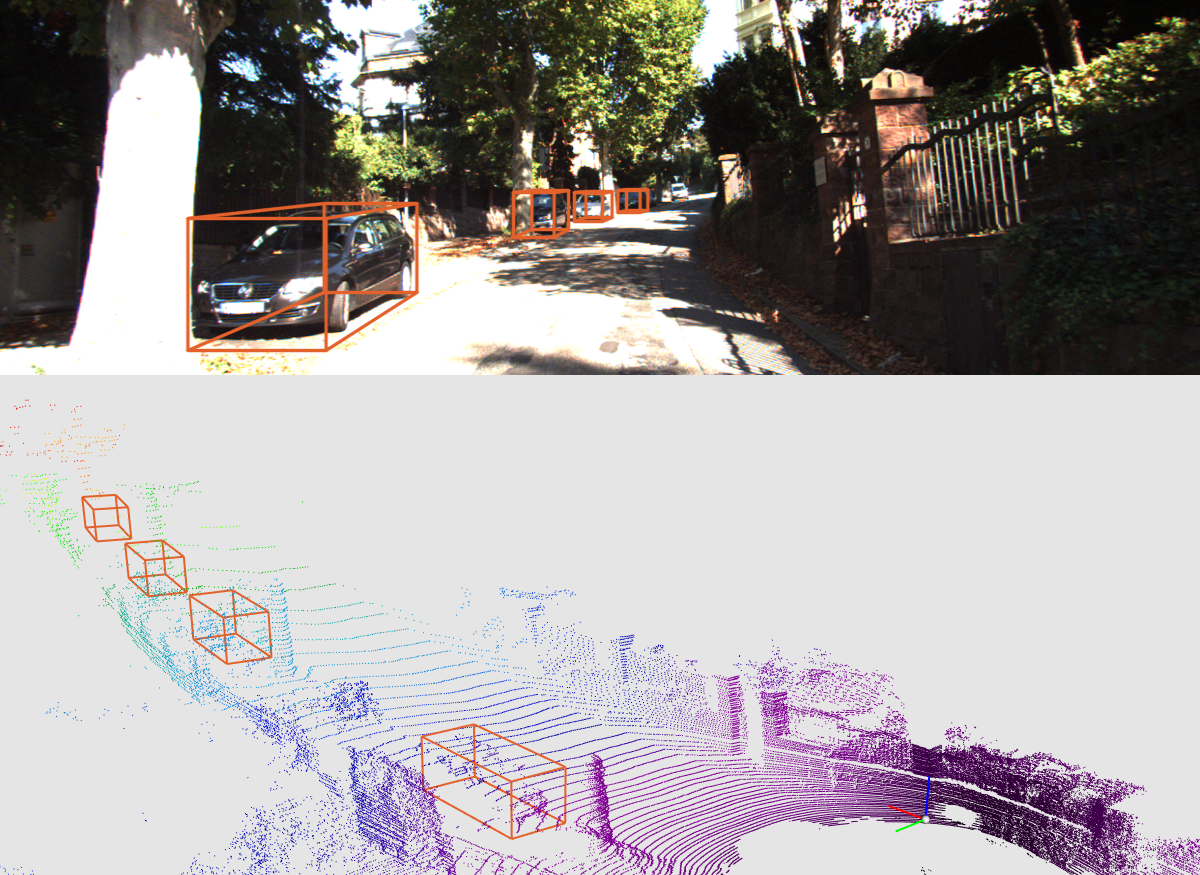}
}

\end{center}
  \caption{
Qualitative results of our method for multi-class 3D object detection. We use orange box for cars, purple box for pedestrians, and green box for cyclists. All illustrated images are from the KITTI \emph{test} set. Zoom in on the images for more details.}
\label{fig: test-1}
\end{figure*}
\begin{figure*}[!t]
\begin{center}
\subfigure{
\includegraphics[width=0.49\linewidth]{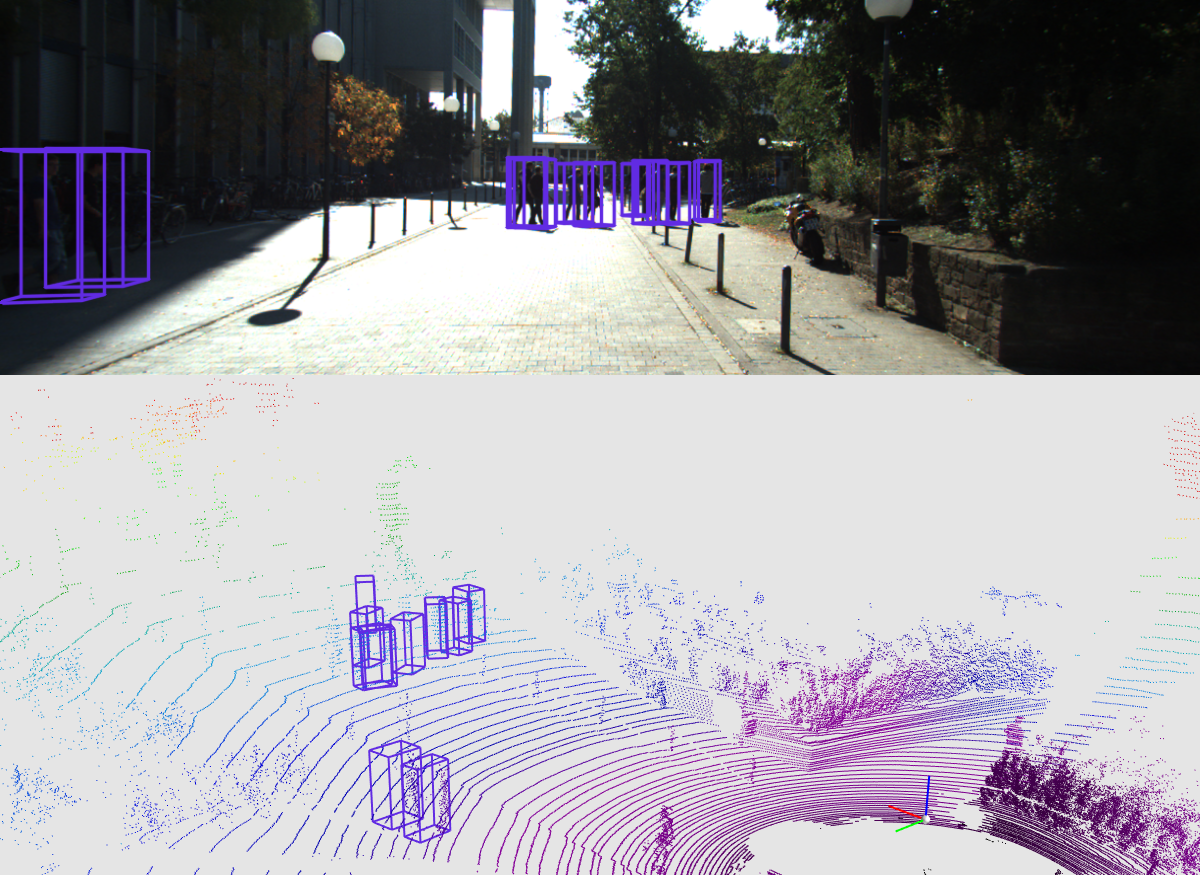}
}\hspace{-0.1 in}
\subfigure{
\includegraphics[width=0.49\linewidth]{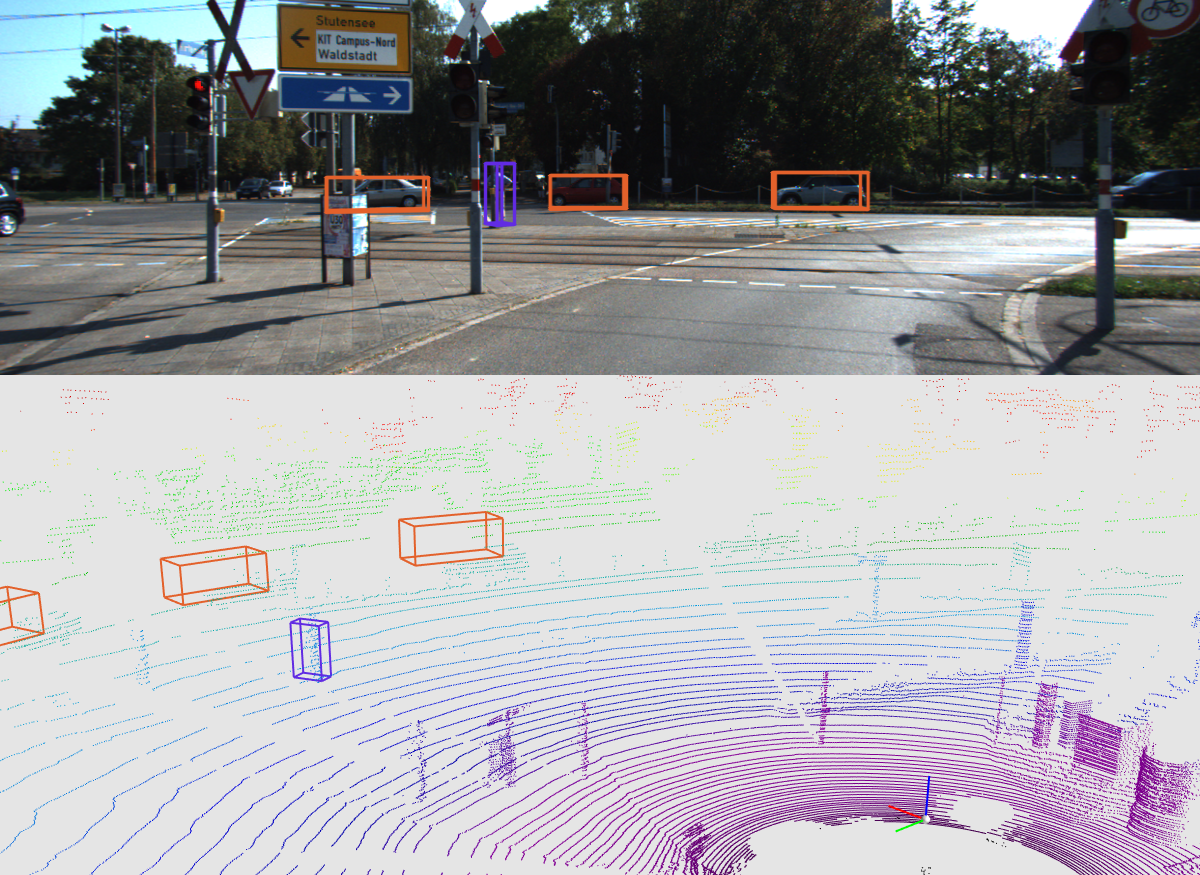}
}

\subfigure{
\includegraphics[width=0.49\linewidth]{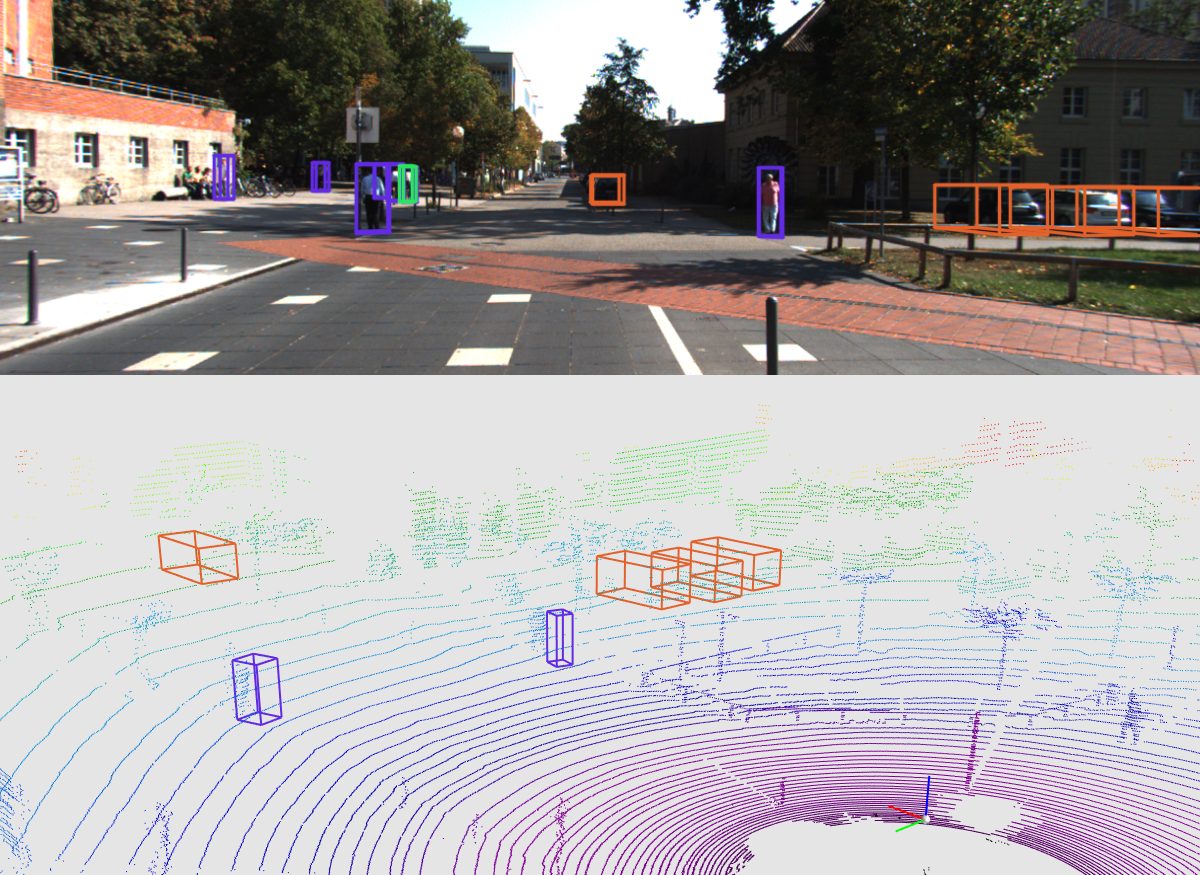}
}\hspace{-0.1 in}
\subfigure{
\includegraphics[width=0.49\linewidth]{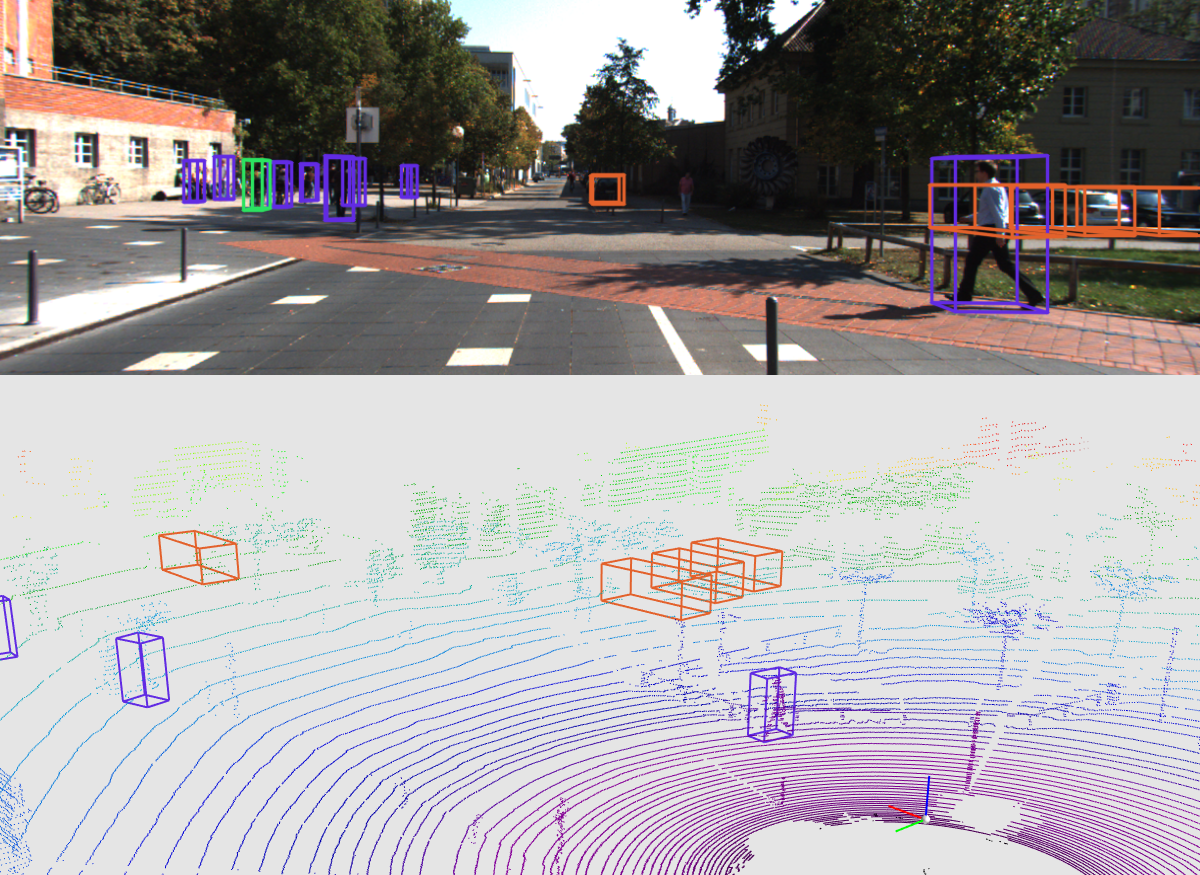}
}

\subfigure{
\includegraphics[width=0.49\linewidth]{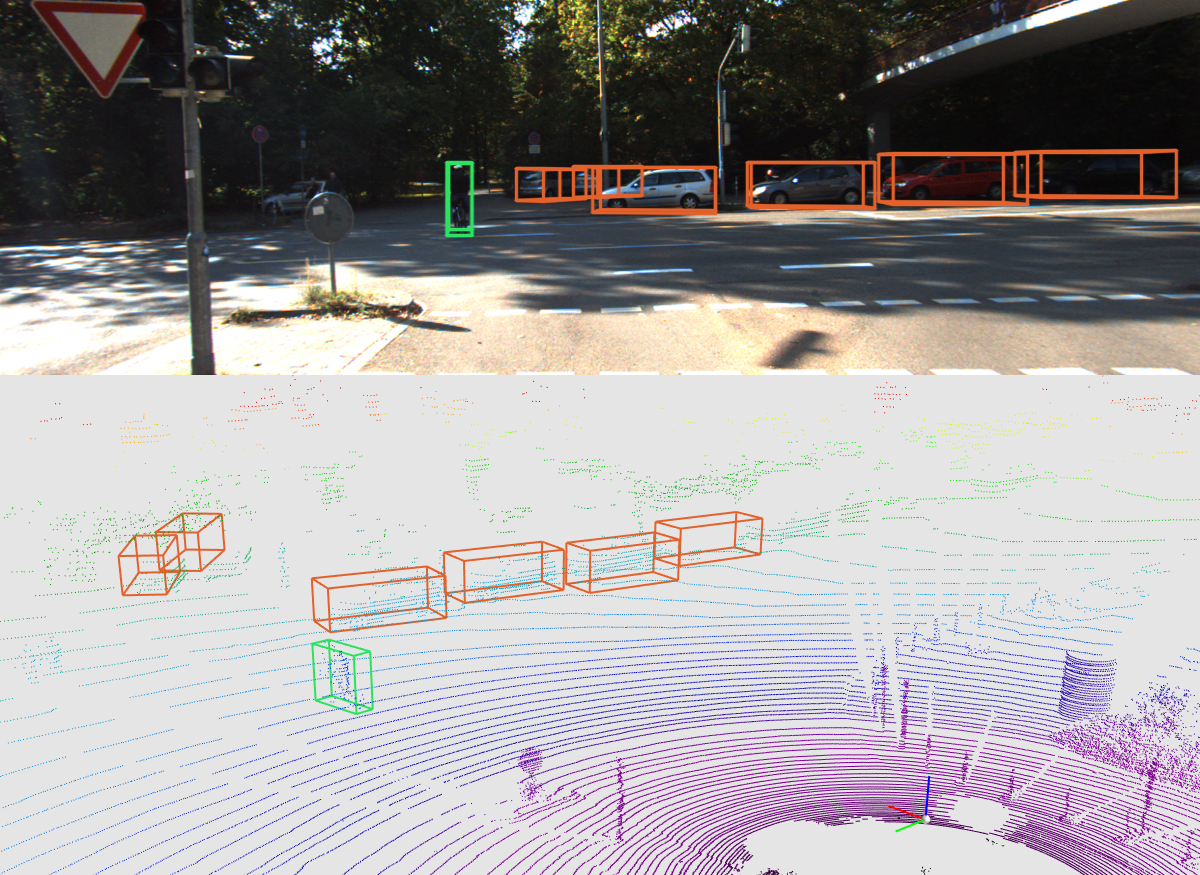}
}\hspace{-0.1 in}
\subfigure{
\includegraphics[width=0.49\linewidth]{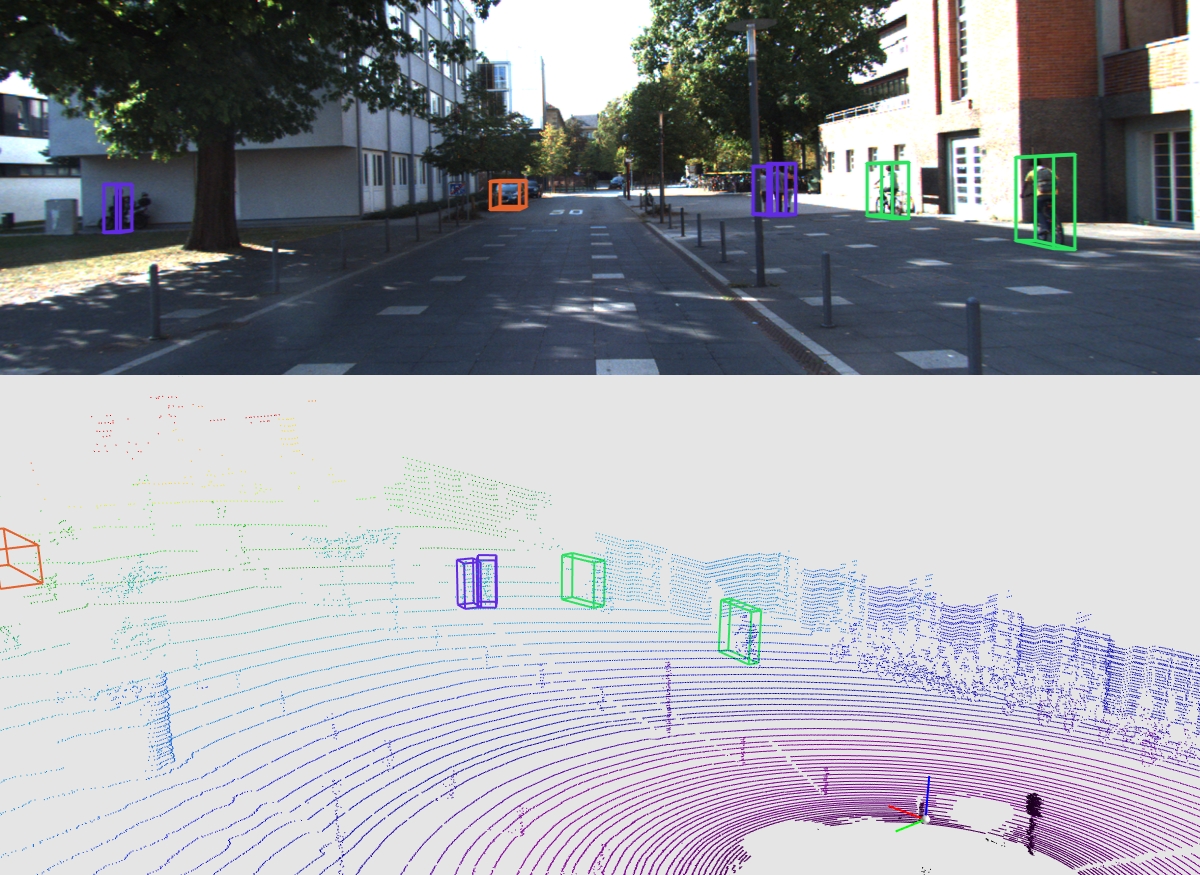}
}

\end{center}
  \caption{
Qualitative results of our method for multi-class 3D object detection. We use orange box for cars, purple box for pedestrians, and green box for cyclists. All illustrated images are from the KITTI \emph{test} set. Zoom in on the images for more details.}
\label{fig: test-2}
\end{figure*}

Figure~\ref{fig: test-1} and~\ref{fig: test-2} show additional visualization of the prediction results on KITTI 3D raw data in both the image plane and the LiDAR coordinate system, respectively. 
We use orange box, purple box, and green box for car, pedestrian, and cyclist, respectively.
Our approach is able to accurately localize the different-depth 3D objects.


\clearpage

{\small
\bibliographystyle{ieee_fullname}
\bibliography{egbib}

\begin{thebibliography}{10}\itemsep=-1pt

\bibitem{keypoint}
Ivan Barabanau, Alexey Artemov, Evgeny Burnaev, and Vyacheslav Murashkin.
\newblock Monocular 3d object detection via geometric reasoning on keypoints.
\newblock In {\em VISIGRAPP}, 2020.

\bibitem{brahmbhatt2018geometry}
Samarth Brahmbhatt, Jinwei Gu, Kihwan Kim, James Hays, and Jan Kautz.
\newblock Geometry-aware learning of maps for camera localization.
\newblock In {\em CVPR}, 2018.

\bibitem{m3drpn}
Garrick Brazil and Xiaoming Liu.
\newblock M3d-rpn: Monocular 3d region proposal network for object detection.
\newblock In {\em ICCV}, 2019.

\bibitem{kinematic3d}
Garrick Brazil, Gerard Pons-Moll, Xiaoming Liu, and Bernt Schiele.
\newblock Kinematic 3d object detection in monocular video.
\newblock In {\em ECCV}, 2020.

\bibitem{decoupled}
Yingjie Cai, Buyu Li, Zeyu Jiao, Hongsheng Li, Xingyu Zeng, and Xiaogang Wang.
\newblock Monocular 3d object detection with decoupled structured polygon estimation and height-guided depth estimation.
\newblock In {\em AAAI}, 2020.

\bibitem{deepmanta}
Florian Chabot, Mohamed Chaouch, Jaonary Rabarisoa, C{\'e}line Teuliere, and Thierry Chateau.
\newblock Deep manta: A coarse-to-fine many-task network for joint 2d and 3d vehicle analysis from monocular image.
\newblock In {\em CVPR}, 2017.

\bibitem{chen20153d}
Xiaozhi Chen, Kaustav Kundu, Yukun Zhu, Andrew~G Berneshawi, Huimin Ma, Sanja Fidler, and Raquel Urtasun.
\newblock 3d object proposals for accurate object class detection.
\newblock In {\em NIPS}, 2015.

\bibitem{fast}
Yilun Chen, Shu Liu, Xiaoyong Shen, and Jiaya Jia.
\newblock Fast point r-cnn.
\newblock In {\em ICCV}, 2019.

\bibitem{monopair}
Yongjian Chen, Lei Tai, Kai Sun, and Mingyang Li.
\newblock Monopair: Monocular 3d object detection using pairwise spatial relationships.
\newblock In {\em CVPR}, 2020.

\bibitem{d4lcn}
Mingyu Ding, Yuqi Huo, Hongwei Yi, Zhe Wang, Jianping Shi, Zhiwu Lu, and Ping Luo.
\newblock Learning depth-guided convolutions for monocular 3d object detection.
\newblock In {\em CVPR}, 2020.

\bibitem{kitti}
Andreas Geiger, Philip Lenz, and Raquel Urtasun.
\newblock Are we ready for autonomous driving? the kitti vision benchmark suite.
\newblock In {\em CVPR}, 2012.

\bibitem{hoiem2008putting}
Derek Hoiem, Alexei~A Efros, and Martial Hebert.
\newblock Putting objects in perspective.
\newblock {\em IJCV}, 2008.

\bibitem{ss3d}
Eskil J{\"{o}}rgensen, Christopher Zach, and Fredrik Kahl.
\newblock Monocular 3d object detection and box fitting trained end-to-end using intersection-over-union loss.
\newblock {\em CoRR}, abs/1906.08070, 2019.

\bibitem{uncertainty}
Alex~Guy Kendall.
\newblock {\em Geometry and uncertainty in deep learning for computer vision}.
\newblock PhD thesis, University of Cambridge, 2019.

\bibitem{monopsr}
Jason Ku, Alex~D Pon, and Steven~L Waslander.
\newblock Monocular 3d object detection leveraging accurate proposals and shape reconstruction.
\newblock In {\em CVPR}, 2019.

\bibitem{groomed-nms}
Abhinav Kumar, Garrick Brazil, and Xiaoming Liu.
\newblock Groomed-nms: Grouped mathematically differentiable nms for monocular 3d object detection.
\newblock In {\em CVPR}, 2021.

\bibitem{pointpillars}
Alex~H Lang, Sourabh Vora, Holger Caesar, Lubing Zhou, Jiong Yang, and Oscar Beijbom.
\newblock Pointpillars: Fast encoders for object detection from point clouds.
\newblock In {\em CVPR}, 2019.

\bibitem{cornernet}
Hei Law and Jia Deng.
\newblock Cornernet: Detecting objects as paired keypoints.
\newblock In {\em ECCV}, 2018.

\bibitem{gs3d}
Buyu Li, Wanli Ouyang, Lu Sheng, Xingyu Zeng, and Xiaogang Wang.
\newblock Gs3d: An efficient 3d object detection framework for autonomous driving.
\newblock In {\em CVPR}, 2019.

\bibitem{stereorcnn}
Peiliang Li, Xiaozhi Chen, and Shaojie Shen.
\newblock Stereo r-cnn based 3d object detection for autonomous driving.
\newblock In {\em CVPR}, 2019.

\bibitem{rtm3d}
Peixuan Li, Huaici Zhao, Pengfei Liu, and Feidao Cao.
\newblock {RTM3D:} real-time monocular 3d detection from object keypoints for autonomous driving.
\newblock In {\em ECCV}, 2020.

\bibitem{focalloss}
Tsung-Yi Lin, Priya Goyal, Ross Girshick, Kaiming He, and Piotr Doll{\'a}r.
\newblock Focal loss for dense object detection.
\newblock In {\em ICCV}, 2017.

\bibitem{liu2020reinforced}
Lijie Liu, Chufan Wu, Jiwen Lu, Lingxi Xie, Jie Zhou, and Qi Tian.
\newblock Reinforced axial refinement network for monocular 3d object detection.
\newblock In {\em ECCV}, 2020.

\bibitem{smoke}
Zechen Liu, Zizhang Wu, and Roland T{\'o}th.
\newblock Smoke: Single-stage monocular 3d object detection via keypoint estimation.
\newblock In {\em CVPR}, 2020.

\bibitem{patchnet}
Xinzhu Ma, Shinan Liu, Zhiyi Xia, Hongwen Zhang, Xingyu Zeng, and Wanli Ouyang.
\newblock Rethinking pseudo-lidar representation.
\newblock In {\em ECCV}, 2020.

\bibitem{am3d}
Xinzhu Ma, Zhihui Wang, Haojie Li, Pengbo Zhang, Wanli Ouyang, and Xin Fan.
\newblock Accurate monocular 3d object detection via color-embedded 3d reconstruction for autonomous driving.
\newblock In {\em ICCV}, 2019.

\bibitem{monodle}
Xinzhu Ma, Yinmin Zhang, Dan Xu, Dongzhan Zhou, Shuai Yi, Haojie Li, and Wanli Ouyang.
\newblock Delving into localization errors for monocular 3d object detection.
\newblock In {\em CVPR}, 2021.

\bibitem{roi10d}
Fabian Manhardt, Wadim Kehl, and Adrien Gaidon.
\newblock Roi-10d: Monocular lifting of 2d detection to 6d pose and metric shape.
\newblock In {\em CVPR}, 2019.

\bibitem{deep3dbox}
Arsalan Mousavian, Dragomir Anguelov, John Flynn, and Jana Kosecka.
\newblock 3d bounding box estimation using deep learning and geometry.
\newblock In {\em CVPR}, 2017.

\bibitem{fpoint}
Charles~R Qi, Wei Liu, Chenxia Wu, Hao Su, and Leonidas~J Guibas.
\newblock Frustum pointnets for 3d object detection from rgb-d data.
\newblock In {\em CVPR}, 2018.

\bibitem{monogrnet}
Zengyi Qin, Jinglu Wang, and Yan Lu.
\newblock Monogrnet: A geometric reasoning network for monocular 3d object localization.
\newblock In {\em AAAI}, 2019.

\bibitem{CaDDN}
Cody Reading, Ali Harakeh, Julia Chae, and Steven~L. Waslander.
\newblock Categorical depth distribution network for monocular 3d object detection.
\newblock {\em CVPR}, 2021.

\bibitem{categorical}
Cody Reading, Ali Harakeh, Julia Chae, and Steven~L Waslander.
\newblock Categorical depth distribution network for monocular 3d object detection.
\newblock In {\em CVPR}, 2021.

\bibitem{rhodin2018unsupervised}
Helge Rhodin, Mathieu Salzmann, and Pascal Fua.
\newblock Unsupervised geometry-aware representation for 3d human pose estimation.
\newblock In {\em ECCV}, 2018.

\bibitem{oftnet}
Thomas Roddick, Alex Kendall, and Roberto Cipolla.
\newblock Orthographic feature transform for monocular 3d object detection.
\newblock In {\em BMVC}, 2019.

\bibitem{Sheng2019Unsupervised}
Lu Sheng, Dan Xu, Wanli Ouyang, and Xiaogang Wang.
\newblock Unsupervised collaborative learning of keyframe detection and visual odometry towards monocular deep slam.
\newblock In {\em ICCV}, 2019.

\bibitem{pv-rcnn}
Shaoshuai Shi, Chaoxu Guo, Li Jiang, Zhe Wang, Jianping Shi, Xiaogang Wang, and Hongsheng Li.
\newblock Pv-rcnn: Point-voxel feature set abstraction for 3d object detection.
\newblock In {\em CVPR}, 2020.

\bibitem{shi2020points}
Shaoshuai Shi, Zhe Wang, Jianping Shi, Xiaogang Wang, and Hongsheng Li.
\newblock From points to parts: 3d object detection from point cloud with part-aware and part-aggregation network.
\newblock {\em TPAMI}, 2020.

\bibitem{ur3d}
Xuepeng Shi, Zhixiang Chen, and Tae-Kyun Kim.
\newblock Distance-normalized unified representation for monocular 3d object detection.
\newblock In {\em ECCV}, 2020.

\bibitem{monodis}
Andrea Simonelli, Samuel~Rota Bulo, Lorenzo Porzi, Manuel L{\'o}pez-Antequera, and Peter Kontschieder.
\newblock Disentangling monocular 3d object detection.
\newblock In {\em ICCV}, 2019.

\bibitem{MoVi-3D}
Andrea Simonelli, Samuel~Rota Bul{\`{o}}, Lorenzo Porzi, Elisa Ricci, and Peter Kontschieder.
\newblock Towards generalization across depth for monocular 3d object detection.
\newblock In {\em ECCV}, 2020.

\bibitem{fcos}
Zhi Tian, Chunhua Shen, Hao Chen, and Tong He.
\newblock Fcos: Fully convolutional one-stage object detection.
\newblock In {\em ICCV}, 2019.

\bibitem{ddmp-3d}
Li Wang, Liang Du, Xiaoqing Ye, Yanwei Fu, Guodong Guo, Xiangyang Xue, Jianfeng Feng, and Li Zhang.
\newblock Depth-conditioned dynamic message propagation for monocular 3d object detection.
\newblock In {\em CVPR}, 2021.

\bibitem{xu2018multi}
Bin Xu and Zhenzhong Chen.
\newblock Multi-level fusion based 3d object detection from monocular images.
\newblock In {\em CVPR}, 2018.

\bibitem{XuMoving}
Dan Xu, Andrea Vedaldi, and João F.~Henriques.
\newblock Moving slam: Fully unsupervised deep learning in non-rigid scenes.
\newblock In {\em IROS}, 2021.

\bibitem{xu2019geometry}
Dan Xu, Weidi Xie, and Andrew Zisserman.
\newblock Geometry-aware video object detection for static cameras.
\newblock In {\em BMVC}, 2019.

\bibitem{reppoints}
Ze Yang, Shaohui Liu, Han Hu, Liwei Wang, and Stephen Lin.
\newblock Reppoints: Point set representation for object detection.
\newblock In {\em ICCV}, 2019.

\bibitem{DA-3Ddet}
Xiaoqing Ye, Liang Du, Yifeng Shi, Yingying Li, Xiao Tan, Jianfeng Feng, Errui Ding, and Shilei Wen.
\newblock Monocular 3d object detection via feature domain adaptation.
\newblock In {\em ECCV}, 2020.

\bibitem{dla}
Fisher Yu, Dequan Wang, Evan Shelhamer, and Trevor Darrell.
\newblock Deep layer aggregation.
\newblock In {\em CVPR}, 2018.

\bibitem{centernet}
Xingyi Zhou, Dequan Wang, and Philipp Kr{\"a}henb{\"u}hl.
\newblock Objects as points.
\newblock In {\em arXiv preprint arXiv:1904.07850}, 2019.

\end{thebibliography}
}

\end{document}